\documentclass[a4paper,fleqn]{cas-sc}

\usepackage{amssymb}
\usepackage{amsmath}
\usepackage{booktabs}
\usepackage{longtable}
\usepackage{lineno}
\usepackage{siunitx}
\usepackage{multirow}
\usepackage{makecell}  
\usepackage{setspace}
\usepackage{caption}
\usepackage{url}
\usepackage{orcidlink}
\usepackage{float}

\usepackage[round]{natbib}
\usepackage{xcolor}
\usepackage{listings}

\hypersetup{
  colorlinks,
  citecolor=cyan,
  linkcolor=cyan,
  urlcolor=cyan}

\begin{document}

\shorttitle{PlantXpert: a Multimodal LLM Benchmark for Plant Phenotyping}
\shortauthors{Wu et al. (2026)}



\title [mode=title]{From Scientific Crop Imagery to Agronomic Reasoning: A Multimodal LLM Benchmark for Plant Phenotyping}


\author[aff0, label2]{Yu Wu\orcidlink{0009-0004-2534-6339}} 
\author[aff1, label2]{Guangzeng Han\orcidlink{0009-0006-8355-2768}}
\author[aff1]{Ibra Niang Niang}
\affiliation[aff0]{organization={Electrical Engineering and Computer Science, University of Missouri},
            city={Columbia},
            postcode={65201},
            state={MO},
            country={United States}}
\affiliation[aff1]{organization={Computer Science, University of Memphis},
            city={Memphis},
            postcode={38111},
            state={TN},
            country={United States}}
\author[aff2]{Francia Ravelombola\orcidlink{0009-0006-1447-2393}}
\author[aff2]{Maiara Oliveira\orcidlink{0000-0003-4283-9763}}
\affiliation[aff2]{organization={Fisher Delta Research Extension and Education Center, University of Missouri},
            city={Portageville},
            postcode={63873},
            state={MO},
            country={United States}}
\author[aff3]{Jason Davis}
\affiliation[aff3]{organization={University of Arkansas Division of Agriculture},
            city={Batesville},
            postcode={72501},
            state={AR},
            country={United States}}
\author[aff4]{Dong Chen\orcidlink{0000-0002-0551-5084}}
\affiliation[aff4]{organization={Agricultural \& Biological Engineering, Mississippi State University},
            city={Starkville},
            postcode={39762},
            state={MS},
            country={United States}}
\author[aff2]{Feng Lin\corref{cor1}}
\author[aff0]{Xiaolei Huang\corref{cor1}\orcidlink{0000-0003-0478-8715}}[orcid=0000-0003-0478-8715]
\cortext[cor1]{Corresponding authors. fenglin104@gmail.com (Lin), xiaolei.huang@missouri.edu (Huang)}
\fntext[label2]{Equal Contributions.}


\begin{abstract}
High-throughput, effective and comprehensive phenotyping is a critical prerequisite for crop improvement. 
While phenotypic analysis was traditionally performed manually, recent advances in multimodal foundation models, especially in vision-language models (VLMs), have offered new opportunities for automated plant phenotyping. 
However, plant science remains particularly challenging for general-purpose VLMs because it requires domain-specific knowledge, fine-grained visual interpretation, and complex agronomic reasoning.
To address this gap, we introduce PlantXpert, an evidence-grounded multimodal reasoning benchmark for plant phenotyping across four major crops: soybean, cotton, rice, and corn. 
Our benchmark provides a structured and reproducible framework for agronomic adaptation of VLMs, and enables controlled comparison between base models and their domain-adapted counterparts. 
We constructed a dataset comprising 545 digital images and 4{,}351 benchmark samples spanning key agronomic domains including disease, pest control, weed management and yield. 
The benchmark can assess diverse capabilities including visual expertise, quantitative reasoning, and multi-step agronomic reasoning. 
We trained and evaluated a total of 11 state-of-the-art VLMs.
Fine-tuning on our dataset improves overall accuracy for all models, with many achieving around 70\% accuracy, and improvements on 9 of 11 remaining statistically significant after Holm correction.
We also identify diminishing returns from model scaling, uneven performance across crop subsets, and persistent weaknesses in quantitative reasoning and specialized image analysis.
These limitations show that general multimodal competence alone is insufficient for reliable agronomic reasoning and demonstrate the value of domain-specific adaptation.
We hope PlantXpert can serve as a foundation for advancing evidence-grounded multimodal reasoning in plant science.


\end{abstract}



\maketitle
\doublespacing



\section{Introduction}

Improving cultivated crops remains a central objective, yet an unsolved challenge in plant science.
For example, soybean is a primary source of high-quality protein and oil \citep{anderson_soybean_2019}, cotton is cultivated across diverse environmental conditions for its fiber, oil, and protein \citep{waghmare_cotton_2022}, rice is vital for the food security of more than half the world’s population \citep{mohidem_rice_2022}, while corn is widely used as animal feed and fuel bioethanol crop \citep{GARCIALARA_corn_2019}. 
Improving the genetics of these crops requires accurate interpretations of plant traits. 
This underlines the importance of high-throughput, effective and comprehensive trait data \citep{mir_high-throughput_2019}. 
However, transforming these data into useful phenotypic insight requires robust and automated analysis. 
Recently, foundation models have emerged as a promising tool for plant phenotyping tasks \citep{li_foundation_2024}. 
These models are pretrained on large-scale data and can be adapted to a wide range of downstream tasks with minimal supervision \citep{bommasani_opportunities_2022}. 
In particular, some multimodal foundation models combine vision encoders with existing language models, commonly known as vision-language models (VLMs). 
These models are capable of integrating visual and contextual information to support in-depth reasoning, providing insights that are unattainable through traditional techniques \citep{Sapkota_multimodal_2025}. 
In addition, their cross-domain generalization abilities make them particularly suitable for phenotypic analysis, where variation in plant architecture and environment is very common \citep{wang_artificial_2025}. 
Despite the promise of VLMs, their application in plant science remains constrained by data-related challenges, including heterogeneous data sources, limited expert annotations, and insufficient ground truth data for model validation \citep{williamson_data_2023}. 
These limitations highlight the need for dedicated benchmarking frameworks that enable systematic and reproducible evaluation of VLMs in realistic agricultural settings.

Earlier plant image datasets, such as PlantVillage \citep{mohanty_using_2016}, PlantDoc \citep{singh_plantdoc_2020}, and FruitNet \citep{meshram_fruitnet_2022}, have been instrumental in the development of specialized computer vision models in plant science. 
However, they exhibit limited diversity with respect to image types, environments, or imaging conditions. 
More importantly, these datasets lack comprehensive annotations or metadata that provide information on phenological stages and other agronomic contexts. 
As a result, they are well suited for tasks such as supervised disease classification or canopy segmentation, but remain insufficient to support more complex phenotypic reasoning required by VLMs. To address this gap, several works have made significant progress in developing agricultural Visual-Question Answering (VQA) datasets for VLM training, including Agri-LLaVA \citep{wang_agri-llava_2024}, AgroInstruct \citep{awais_agrogpt_2025}, and Agri-342k \citep{yang_agrigpt_2025}. 
In addition, some datasets, such as CDDM \citep{liu_multimodal_2025} and LeafNet \citep{quoc_leafnet_2026}, focus specifically on training VLMs for disease diagnosis. 
These datasets often prioritize perception-level task adaptation, but are less explicitly designed for controlled evaluation of evidence-grounded phenotypic reasoning over scientific imagery.
Meanwhile, several systematic benchmark frameworks tailored for the evaluation of existing VLMs have also been published. 
AgriBench is a hierarchical benchmark to evaluate multimodal large language models (LLM) for agricultural applications \citep{zhou_agribench_2025}, including tasks ranging from basic crop detection to high stakes decision making. 
AgEval presents a more task-specific framework, focusing on the potential of VLMs for plant stress phenotyping \citep{Arshad_WACV_2025}. 
AgroBench is a comprehensive benchmark with more than 4{,}000 expert-validated question-answer (QA) pairs across multiple identification and management tasks for more than 200 crops \citep{Shinoda_ICCV_2025}. 
And AgriEval \citep{yan_agrieval_2026} provides another large-scale benchmark with around 17{,}000 QA pairs covering six major agricultural categories, from plant production to forestry. 
Despite these advances, most existing benchmarks are evaluated with zero-shot or few-shot settings only, with limited support for standardized supervised adaptation or controlled fine-tuning.  
As a result, lack of knowledge remains a critical problem, accounting for 51.92\% of errors in Agrobench \citep{Shinoda_ICCV_2025}.
It is difficult to assess how effectively VLMs acquire domain-specific knowledge and improve multimodal reasoning tasks that require more than surface-level recognition.
In addition, many benchmarks provide limited structured annotation, such as agronomic domains and challenge types, restricting fine-grained analysis of model behavior across diverse plant science contexts. 
More importantly, benchmark samples are often not explicitly constructed around scientific visual evidence and its associated textual interpretation, making it difficult to evaluate whether models can perform evidence-grounded reasoning in realistic phenotyping scenarios. 
These limitations call for benchmark frameworks that not only evaluate multimodal reasoning capabilities but also support controlled agronomic adaptation of VLMs, enabling more reliable and domain-aware applications in plant phenotyping.

To address these challenges, we propose \textbf{PlantXpert}, an evidence-grounded multimodal reasoning benchmark for plant phenotyping, centered on four widely grown crops: soybean, cotton, rice, and corn. 
Rather than serving as a conventional VQA dataset, our benchmark is designed to evaluate whether VLMs can interpret scientific agricultural figures by grounding their decisions in visual evidence and associated agronomic context. 
In addition, since it provides a structured and reproducible evaluation framework in both zero-shot and fine-tuned settings, PlantXpert enables controlled comparison between base models and their fine-tuned counterparts. 
Unlike prior benchmarks that focus primarily on zero-shot or few-shot evaluation, our framework explicitly supports supervised adaptation, allowing us to study how fine-tuning affects domain-specific reasoning in plant science settings. 
The dataset, consisting of more than 4{,}000 QA pairs, is divided into train and test splits, with the test set annotated by human experts for validation. 
PlantXpert covers not only well-studied tasks such as plant disease identification, but also more reasoning-intensive scenarios including phenotype-based causal temporal and quantitative interpretation. 
For example, we prompt models to infer disease progression by connecting visible symptoms to development stages. 
Thus, our benchmark moves beyond surface-level recognition and instead emphasizes evidence-grounded reasoning.  
In addition, PlantXpert incorporates comprehensive structured metadata for each data sample, including agronomic domain, primary challenge type, and detailed explanations of visual evidence and reasoning difficulty. 
This structure enables fine-grained evaluation of model performance across different reasoning categories, such as spatial interpretation, causal inference, quantitative estimation, and expert-level biological understanding. 
Furthermore, by linking each sample to its original scientific source, such as figure captions and related contexts, the benchmark preserves data provenance and supports more interpretable and reliable evaluation. 
Through this design, PlantXpert goes beyond conventional VQA datasets, allowing not only performance measurement but also systematic diagnosis of model limitations in evidence-grounded agronomic reasoning.

\begin{table}[t]
\centering
\caption{Comparison of existing plant science benchmarks with our proposed framework across key capabilities.}
\label{tab:benchmark_comparison}
\resizebox{0.95\linewidth}{!}{
\begin{tabular}{lccccc}

\textbf{Study} & \textbf{Multimodal} & \textbf{Phenotypic Reasoning} & \textbf{Expert Annotation} & \textbf{Evidence-Grounded} & \textbf{Benchmark-Training Integration} \\
\midrule
AgriBench~\citep{zhou_agribench_2025} & \checkmark & $\times$ & $\times$ & $\times$ & $\times$ \\
AgEval~\citep{Arshad_WACV_2025} & \checkmark & \checkmark & $\times$ & $\times$ & $\times$ \\
AgroBench~\citep{Shinoda_ICCV_2025} & \checkmark & \checkmark & \checkmark & $\times$ & $\times$ \\
AgriEval~\citep{yan_agrieval_2026} & $\times$ & \checkmark & \checkmark & $\times$ & $\times$ \\
\midrule
PlantXpert (Ours) & \checkmark & \checkmark & \checkmark & \checkmark & \checkmark \\

\end{tabular}
}
\end{table}

\section{PlantXpert: An Evidence-grounded Reasoning Benchmark for Plant Phenotyping}

Plant phenotyping requires models to connect visual observations with biological interpretations and agronomic contexts, while existing studies of LLM reasoning for plant sciences~\citep{zhang2024cmmmu,yue2024mmmu,quoc2026leafnet} often measure surface-level visual recognition without sufficient contextual and scientific grounding.
This study constructs an evidence-grounded multimodal reasoning benchmark centered on plant figures, \textbf{PlantXpert}.
Very few plant phenotyping studies have provided structured annotations and rich agronomic evidence to enhance and evaluate the reasoning capabilities of LLMs, let alone expert-level scientific annotations, while several existing closely related studies~\citep{Arshad_WACV_2025,Shinoda_ICCV_2025,zhou_agribench_2025} in Table~\ref{tab:benchmark_comparison} have validated multimodal LLMs in zero-shot / few-shot testing, differing from our work.
For our study, rather than treating each sample as a direct image-to-text annotation instance, we built it as a grounded reasoning unit supported by scientific visual evidence and corresponding agronomic interpretation, which we extracted from scientific literature. 
This design reflects the nature of plant phenotyping, which often requires more than identifying visible symptoms or objects and instead depends on interpreting developmental stages, stress responses, disease progression, spatial patterns, and other biologically meaningful cues within an agronomic context.
Our benchmark supports not only end-task evaluation beyond zero/few-shot settings, but also fine-grained diagnosis of model behavior and controlled comparison between pretrained and domain-specialized VLMs with rich scientific grounding contexts.

\begin{figure*}[]
\centering
\includegraphics[width=0.95 \textwidth]{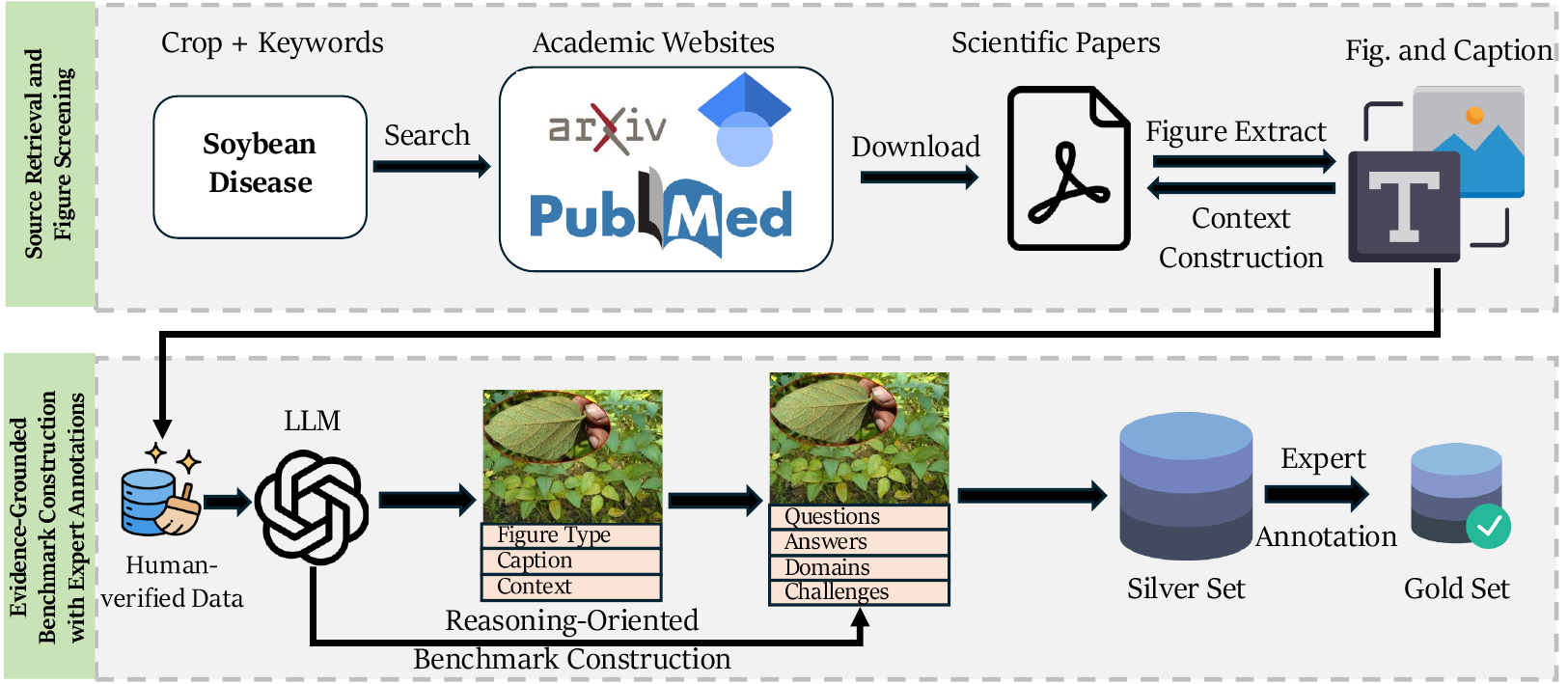} 
\caption{An overview of our source retrieval benchmark construction pipeline.}
\label{fig:overview}
\end{figure*}

We developed PlantXpert based on three design principles, scientific grounding, model reasoning capability, and diagnostic evaluation. 
First, \textit{scientific grounding}: every benchmark sample is tied to a source scientific figure and its associated textual evidence, rather than being generated from the image alone. 
Second, \textit{reasoning-oriented construction}: we use LLMs not only to scale annotation but also to help compose benchmark samples that target higher-level phenotypic reasoning rather than generic visual matching. 
Third, \textit{diagnostic evaluation}: each item is annotated with agronomic domain and primary challenge type, enabling performance analysis beyond aggregate accuracy. Based on these principles, our construction pipeline consists of two major stages: \textit{Source Retrieval and Figure Screening}, and \textit{Evidence-Grounded Benchmark Construction}. An overview of our framework is shown in Figure~\ref{fig:overview}.

\subsection{Source Retrieval and Figure Screening}

To construct agronomic and scientific grounding contexts, we retrieved scientific articles on soybean, cotton, rice, and corn from publicly accessible venues and open repositories, including arXiv, Google Scholar, Semantic Scholar, and PubMed, supplemented when necessary by authorized-access sources (e.g., citations and references).
Our retrieval was guided by the main phenotyping tasks for these four major crops.
Then, the process kept only the research articles aligned with the core agronomic objectives relevant to plant phenotyping, including disease assessment, pest control, weed management, and crop yield.
We repeated the retrieval process separately for the four crops, covering research studies published between 2016 and 2025 using a set of shared keywords across these target domains in Table~\ref{tab:retrieval_stats}.
The initial process removed broken files, redundant copies of the same PDF files, non-research articles, and non-English articles.
This process finally yielded 1685 soybean papers, 793 cotton papers, and 800 papers each for rice and corn.
Following retrieval, we applied a screening procedure by multiple human experts led by three authors to retain only the papers suitable for benchmark construction.
Specifically, we removed duplicates and near-duplicate papers and filtered out documents irrelevant to the targeted agronomic and plant topics.

For each retained PDF, we extracted figure-caption pairs using PDFigCapX~\citep{li_figure_2019}. 
Figure-caption pairs were identified using explicit figure markers, such as "Fig." or "Figure", which were spatially associated with their corresponding captions and graphical regions identified via layout analysis.
Clearly non-informative graphics, such as logos and decorative elements, were excluded.
To improve the reliability of the benchmark source material, all figure-caption pairs were independently reviewed by two undergraduate annotators with solid backgrounds in digital agriculture.
The two undergraduate annotators were instructed to follow these criteria during screening: 
1) agronomy relevance, meaning that the figure contained crop-related visual evidence such as plants, leaves, field scenes, pests, diseases, or phenotyping measurements, rather than non-agronomic illustrations like application screenshots or model diagrams; 
2) figure-caption consistency, meaning that the caption matched the figure correctly and provided sufficient information to interpret the visual content; 
and 3) visual quality and informativeness, meaning that the figure was visually clear and contained legible, scientifically meaningful information.
The student annotators marked a figure-caption pair as "removed" if they judged that it failed to meet one or more of the criteria above. 
Their results were manually screened again by the corresponding author as an additional quality-control step.  
If both student annotators agreed that a figure-caption pair should be removed, we excluded it from the corpus.
In cases of disagreement, the corresponding author reviewed the sample and made the final decision.
After this filtering and quality-control process, the final figure corpus consists of 305 soybean figures and 80 figures each for cotton, rice, and corn, all with rich scientific contexts, such as captions and paper metadata (e.g., date and source).

\subsection{Evidence-Grounded Benchmark Construction with Expert Annotations}

\paragraph{Evidence Grounding and Context Construction.}
We constructed PlantXpert not from images alone.
Instead, for each retained figure, we built an evidence structure that links the figure, caption, and source-document context to enrich visual content interpretation. 
This design reflects the nature of plant phenotyping in scientific practice, where visual evidence becomes more meaningful when considered together with agronomic descriptions and biological interpretation provided in the source document.
To obtain and format this evidence structure, we provided the image, its caption, and the complete source paper to OpenAI GPT-5-mini \citep{singh_openai_2025} and prompted it to extract the textual passages that were specifically relevant to the figure, particularly those explaining the visual evidence or its agronomic significance.
At this stage, the GPT LLM was instructed not to introduce external knowledge or generate unsupported explanations, but to locate and organize context already present in the original documents. 
To maintain evidence-driven grounding, at least two authors of this work manually verified the extracted context to ensure consistency with the source paper and to remove any hallucinated, irrelevant, or overly broad information.
Through this step, each retained sample was transformed from an isolated image instance into a comprehensive data entry with provenance-grounded evidence.
We further asked GPT-5-mini to assign an agronomy-relevant figure-type label to each sample, such as disease symptoms, pest damage, and crop yields.
These labels were not used as benchmark inputs themselves, but as structural cues to guide the subsequent item construction process. 
In other words, figure types function as intermediate control signals that help align the generated items with the underlying visual and agronomic characteristics of each sample. 
Therefore, this stage moved the benchmark beyond generic image description~\citep{liang2025dynamic} or the traditional question-answering benchmark~\citep{yan_agrieval_2026, Shinoda_ICCV_2025}. 
Instead, it better aligns with the model reasoning demands across plant science research.

\begin{figure*}[]
\centering
\includegraphics[width=0.95 \textwidth]{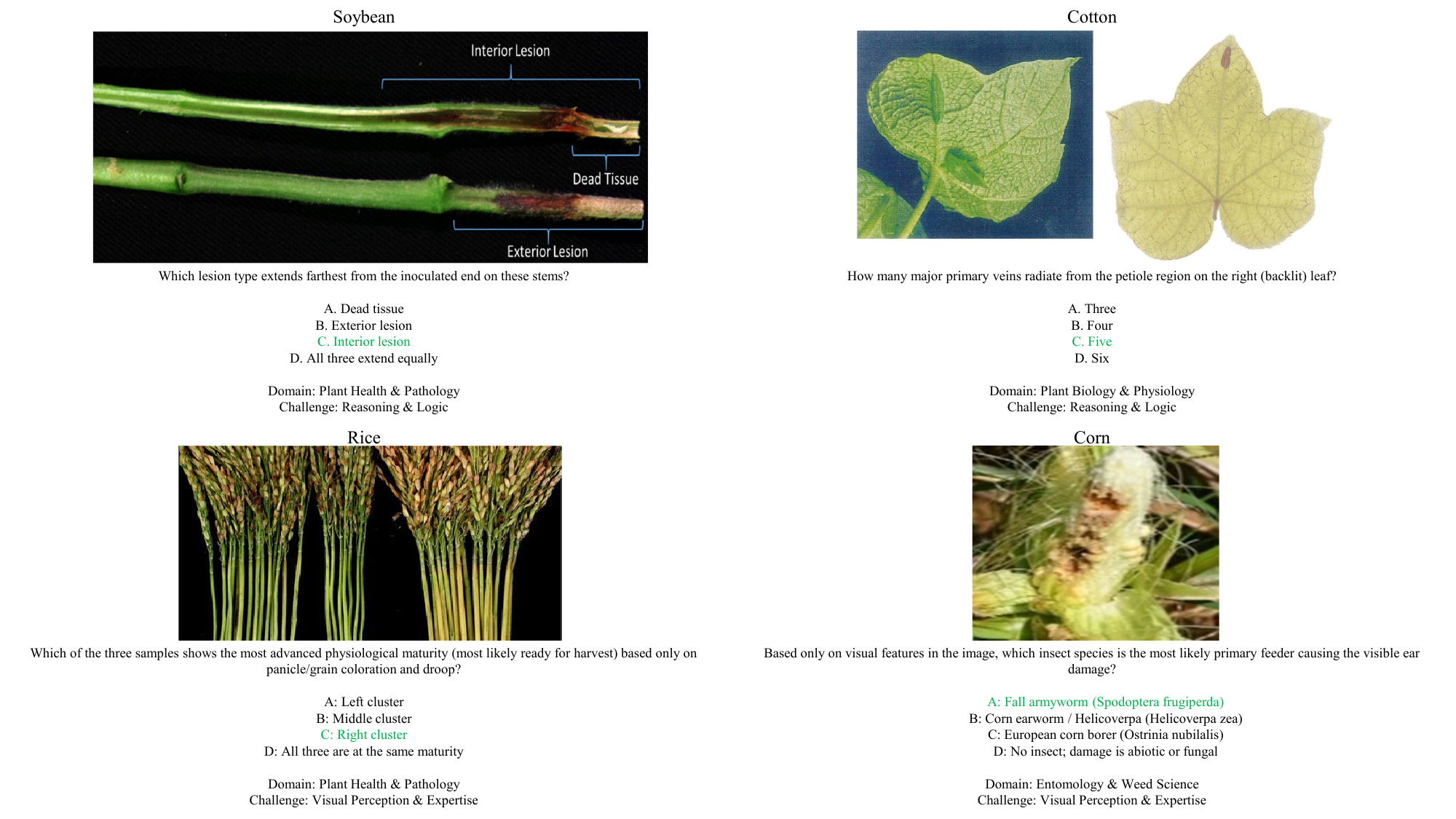} 
\caption{Representative evidence-grounded multiple-choice data samples from PlantXpert. Each sample is annotated with a specific agricultural domain and primary reasoning challenge.}
\label{fig:sample}
\end{figure*}

\paragraph{Reasoning-Oriented Benchmark Construction.}
Based on data entries with grounded evidence, we constructed our benchmark with assistance from the OpenAI API through a constrained, contextualized composition process. 
We leveraged the LLMs beyond an unconstrained data generator to create and enlarge the dataset: we harnessed the reasoning capabilities of the API to integrate visual evidence, caption information, and source-document context to create benchmark samples that more often target non-trivial phenotypic reasoning rather than surface-level recognition. 
Concretely, for each figure, we prompted GPT-5-mini to generate eight multiple-choice examples, each associated with an agronomic domain and a primary challenge type. 
The prompt template is provided in the Appendix~\ref{appendx:prompt}.
During generation, the LLMs had access to the caption and the verified context in addition to the image.
Therefore, the resulting questions-answer pairs are grounded in the caption and source context rather than in image-only speculation.
The generation process yields a benchmark with a diagnostic and inferential structure, rather than a flat collection of multiple-choice prediction examples like the traditional question answering task.
Rather than only asking what phenotype or symptom is visible, the generated data entries aim to probe whether a model can interpret plant traits, identify agronomically meaningful distinctions, estimate quantities, reason about spatial or temporal patterns, and connect visible evidence to biological or management-related implications. 
We leveraged the strong generation capabilities of OpenAI APIs to synthesize higher-order reasoning dimensions during benchmark construction and ground synthetic reasoning data better with real-world plant phenotyping scenarios.
To ensure synthetic data quality, at least two student authors manually screened each generated sample for grounding, correctness, and agronomic relevance.
In total, we obtained 4{,}351 samples, including 2{,}436 for soybean, 635 for rice, and 640 each for cotton and corn. 
Representative examples of our dataset are provided in Figure~\ref{fig:sample}.

\paragraph{Expert Annotation, Adjudication, and Filtering.}
To obtain a high-confidence evaluation set, we constructed an expert-annotated Gold split from the synthetic samples.
Specifically, for each plant, we randomly sampled 20\% of the figures and used questions generated from these figures only to develop annotation pipelines using Label Studio~\citep{Label} for expert review. 
All samples associated with the selected images were included in the Gold split, except for the soybean set, for which 2 questions per figure were used because a larger number of soybean figures were available. 
Each of these samples was subsequently assessed by three of our experts in agronomy, including our co-authors, F. Ravelombola and M. Oliveira, both of whom hold PhD degrees in agronomy. 
For each question, the first two annotators were considered primary experts, while the third served as an adjudicator.
The experts were asked to review and annotate each data entry independently, and the generated answers were hidden from them before the annotations were finished to avoid anchoring effects.
We started with a total of 506 candidate samples, including 122 soybean samples and 128 samples each for cotton, rice, and corn.  
We then assessed inter-annotator agreement among the two primary experts on these valid candidate samples. The primary experts agreed on 120 of the 122 soybean samples (98.4\%; Cohen's $\kappa=0.9672$), 126 of the 128 cotton samples (98.4\%; Cohen's $\kappa=0.9792$), 119 of the 128 rice samples (93.0\%; Cohen's $\kappa=0.9063$) and 117 of the 128 corn samples (91.4\%; Cohen's $\kappa=0.8855$). Overall, agreement was reached on 482 of 506 samples (95.0\%), with Cohen's $\kappa$ of 0.8994, indicating the high quality of both the data entries and annotations.
The final gold label was determined through an adjudication protocol.
If the two primary experts shared the same answer, that answer was adopted directly. 
In the case they disagreed, the third expert's answer was used as a tie breaker. Namely, if an expert's answer matched with the third expert's answer, this answer was accepted. 
If all three answers differed, the sample was discarded. 
Following this final adjudication step, 1 rice and 1 corn samples were discarded and therefore the resulting gold split consisted of 504 human-verified samples, including 122 samples for soybean, 128 samples for cotton, and 127 samples for rice and corn.
To prevent data leakage, all figures used to construct the Gold split and the corresponding questions generated from them were subsequently removed from the original dataset, and the remaining samples were denoted as the Silver split. 
In this way, we were able to construct the Silver and Gold splits that are disjoint at the question and image levels. 
Source-paper-level disjointness is not enforced, meaning that different figures originating from the same publication may occur across the two splits.
We adopt figure-level rather than source-paper-level separation because individual papers contribute highly unequal numbers of usable figures, and a paper-level split could substantially alter the crop and figure-type composition of the benchmark.
The final silver split for training contained 3479 questions in total.
This split strategy serves two complementary purposes: the Silver split supports domain-specific tuning, whereas the Gold split serves as a clean and high-confidence benchmark for evaluation.
This setup allows PlantXpert to provide a controlled training-evaluation setup for fair training and comparison of VLM reasoning capabilities.
We summarize the data statistics of the two splits in Table \ref{tab:domain_challenge_breakdown_silver_gold}.

\begin{table}[t!]
\centering
\small
\setlength{\tabcolsep}{6pt}
\caption{Breakdown of domain and challenge types by crop in the Silver and Gold splits.}
\label{tab:domain_challenge_breakdown_silver_gold}
\begin{tabular}{lcccccccccc}
& \multicolumn{5}{c}{\textbf{Silver}} & \multicolumn{5}{c}{\textbf{Gold}} \\
\cmidrule(lr){2-6}\cmidrule(lr){7-11}
\textbf{Category} &
\textbf{Soybean} &
\textbf{Cotton} &
\textbf{Rice} &
\textbf{Corn} &
\textbf{Total} &
\textbf{Soybean} &
\textbf{Cotton} &
\textbf{Rice} &
\textbf{Corn} &
\textbf{Total}\\
\midrule

\multicolumn{5}{l}{\textbf{Domain}} \\
\midrule
Plant Health \& Pathology       & 1291 & 277 & 108 & 139 & 1815 & 91 & 63 & 22 & 32 & 208\\
Plant Biology \& Physiology     & 149  & 73  & 90  & 82 & 394 & 6 & 18 & 27 & 20 & 71\\
Crop Science \& Agronomy        & 122  & 47  & 114  & 112 & 395 & 7 & 11 & 29 & 30 & 77\\
Remote Sensing \& Imaging       & 123  & 24  & 89  & 70  &  306 & 5 & 23 & 21 & 16 & 65\\
Entomology \& Weed Science      & 65   & 29  & 69  & 64 & 227  & 2 & 8 & 15 & 16 & 41\\
Others                          & 198  & 62  & 37  & 45 & 342 & 11 & 5 & 13 & 13 & 42\\
\midrule

\multicolumn{5}{l}{\textbf{Challenge Type}} \\
\midrule
Reasoning                       & 771  & 179 & 140  & 145 & 1235 & 42 & 45 &  41 & 35 & 163\\
Visual                          & 631  & 174 & 210 & 208 & 1223 & 57 & 44 & 53 & 54 & 208\\
Quantitative                    & 235  & 64  & 115  & 118  & 532 & 17 & 19 & 21 & 30 & 87\\
Others                          & 311  & 95 & 42   & 41  & 489 & 6 & 20 & 12 & 8 & 46\\
\midrule

\textbf{Total}                  & \textbf{1948} & \textbf{512} & \textbf{507} & \textbf{512} & \textbf{3479} &  \textbf{122} & \textbf{128} & \textbf{127} & \textbf{127} & \textbf{504}\\
\end{tabular}
\end{table}

\section{Agronomic VLM Adaptation for Multimodal Reasoning}

Enhancing the agronomic reasoning skills of VLMs is critical to adapting models to plant phenotyping tasks. 
However, very few current studies provide support for agronomic supervision to improve evidence-grounded reasoning.
This study sets up a unified fine-tuning pipeline to enable a reproducible process that improves VLM reasoning on the plant phenotyping task.
Specifically, we present model and supervision formats, prompt instructions, and optimization settings so that performance differences can be attributed to agronomic adaptation rather than inconsistent training conditions.
Our study leverages broad VLM bases with varying sizes and architectures for generalizability demonstrations and reports detailed implementation settings for reproducibility and model adaptation in plant science applications.

\paragraph{Models.}
Agronomic reasoning requires models with diverse architectures and scales to fit varying needs, from cloud deployment to edge devices, yet few plant phenotyping studies evaluate such model diversity under a unified adaptation and reasoning benchmark.
Our study goes beyond single-model reporting and presents a unified model training and comparison pipeline: multiple state-of-the-art VLM variants are trained under the same PlantXpert protocol, allowing us to examine how model scale and agronomic fine-tuning jointly influence reasoning performance.
Specifically, our model training and benchmark include 11 publicly available vision-language models under Base and Task-Specific Fine-Tuned settings, including Qwen3-VL-Instruct \citep{bai_qwen3_2025} (2B/4B/8B/30B), Google Gemma3 \citep{gemmateam_gemma3technicalreport_2025} (4B/12B/27B), LLaVA-Next \citep{liu_llavanext_2024} (8B/34B), OpenGVLab InternVL3.5-2B \citep{wang_internvl3_2025}, and IBM Granite-Vision-3.3-2B \citep{team_granite_2025}.
Our evaluated models vary not only in parameter scale, but also in architectural design, including dense and MoE-style language backbones (e.g., Qwen3-VL-30B vs -8B), different vision encoders and vision-language connectors (e.g., ViT in InternVL vs SigLP in Granite), and multimodal integration paradigms.
All models are publicly available on Hugging Face and evaluated under the same benchmark protocol to ensure fair and consistent comparison.
These models are further grouped into 3 subgroups based on their number of parameters: small (fewer than 4B), medium (8-12B), and large (more than 15B).
This setup allows us to analyze the scaling effects in different evaluation schemes.

\paragraph{Prompt instructions.}
A central challenge in multimodal plant reasoning is that current VLMs often rely on shallow answer matching or direct visual-text association, while many plant phenotyping questions require intermediate reasoning that connects observed image patterns to agronomic and biological interpretation. 
Motivated by this gap, we design our prompting strategy to explicitly encourage multi-step reasoning rather than answer-only prediction. 
In contrast to many existing benchmarks that adopt an Answer-Only prompting format, our approach asks the model to decompose the question, inspect the visual evidence, compare the candidate options, and generate an explicit rationale before selecting the final answer. 
Consequently, we follow the Chain-of-Thought prompting strategy \citep{wei_chain_2022}, in which models are instructed to reason through the problems before selecting the final answer.
This design is intended to encourage models to move beyond direct answer matching and instead make their decisions through intermediate reasoning grounded in visual evidence.
All settings begin with a prompt that instructs the models to analyze the image, consider the four candidate options, and finally return exactly one correct final answer. 
The models take the question and the answer choices as the textual input, and the image as the visual input.
They then integrate visual evidence and textual input through multimodal attention mechanisms. 
Based on these contextualized representations, they autoregressively generate the final answer through next-token prediction: 
$P(y \mid x, I) = \prod_{t=1}^{T} P\!\left(y_t \mid y_{<t}, x, I\right),$ where $y$ denotes the generated output sequence, including the rationale and the final answer, $I$ is the image and $x$ is the textual prompt. 
This autoregressive formulation is well suited to our prompting design because it allows the models to generate both intermediate rationales and final answers within a single output sequence.
In particular, we intend to use the rationale to bridge low-level visual cues and higher-level agronomic interpretation, allowing the model to connect visible evidence with biologically meaningful conclusions.
This formulation encourages models to compare all candidate options, organize intermediate reasoning steps, and select the final answer based on visual evidence rather than pattern-matching at the surface-level alone. 

\paragraph{Supervision settings.}
A key challenge in examining and characterizing agronomic VLM reasoning is to distinguish what current VLMs can achieve from general-purpose multimodal pretraining alone from what they can acquire through domain-specific supervision.
We define two controlled supervision settings, Base and Tuned, to measure how agronomic adaptation changes reasoning performance in plant phenotyping.
For the Base setting, we use the original released model checkpoints with no additional task-specific examples provided. 
This allows us to measure the extent to which general-purpose multimodal pretraining transfers to plant phenotyping. 
For the Tuned setting, we further fine-tuned the Base models on our Silver split. 
Under both settings, we subsequently evaluate the models on the non-overlapping Gold split.
This paired setup enables controlled comparison between pretrained and domain-specialized versions of the same model, thereby isolating the effect of agronomic supervision on multimodal reasoning performance.
For model input, we format each sample in the Silver split as a single-image multiple-choice reasoning problem consisting of one scientific figure, one question, and candidate answer options.
Other original and raw contexts, such as structured metadata, figure-type labels, and source-paper paragraphs, are not exposed during fine-tuning to prevent information leakage.
We use LLaMA-Factory \citep{zheng-llamafactory-2024} for Low-Rank Adaptation (LoRA; \citealp{hu_lora_2022})-based supervised fine-tuning (SFT) on our Silver split.
LoRA is a parameter-efficient fine-tuning method that adapts a pretrained model by inserting small trainable low-rank matrices into selected layers. 
We intentionally adopt a lightweight and standardized parameter-efficient tuning strategy rather than custom model-specific optimization, so that performance differences more cleanly reflect the effect of agronomic supervision rather than architecture-specific engineering choice.
The supervised settings allow us to assess how controlled domain adaptation improves evidence-grounded agronomic reasoning beyond the capabilities of the original pretrained models.

\paragraph{Implementation.}
A reproducible implementation setting is important to ensure that performance differences reflect model adaptation rather than inconsistencies in training infrastructure or optimization.
We perform all implementation and fine-tuning on the iTiger GPU cluster~\citep{sharif_2025_ITIGER} with NVIDIA H100 80GB and RTX 6000 48GB GPUs, with GPU allocation varying according to model size. 
We perform optimization using the AdamW optimizer from PyTorch together with a cosine learning-rate scheduler and a warm-up ratio of 0.03.
For the LoRA adaptation, we set the LoRA rank to 16, the LoRA alpha to 32, and the LoRA dropout to 0.05.
We use a per-device batch size of 1 with gradient accumulation over 4 steps, which gives an effective batch size of 4 and a learning rate of $1 \times 10^{-4}$. The only exception is  Qwen3-VL-30B-Instruct \citep{bai_qwen3_2025}, for which we use $1 \times 10^{-5}$.
Under the Tuned setting, we apply three epochs of LoRA-based SFT to each model, and the final checkpoint is saved as our Tuned model for evaluation.
We set the maximum sequence length to 4096 tokens for all models except IBM Granite-Vision-3.3-2B \citep{team_granite_2025}, for which we use 16384. 
The purpose is to accommodate multimodal input and avoid possible truncation during training.
Together, these implementation choices provide a consistent and controlled adaptation setting, allowing us to interpret performance differences more directly in terms of agronomic supervision rather than training variability.

\begin{table*}[t]
\centering
\small
\setlength{\tabcolsep}{5pt}
\sisetup{
  table-number-alignment = center,
  table-format = 2.2,
  detect-weight = true
}

\caption{Accuracy (\%) on the subsets by plant (mean over 3 repeated inference runs) for the Base and Tuned models.
$\Delta$ represents the Base-to-Tuned improvement in overall accuracy, in percentage points.}
\label{tab:vlm_0shot_base_vs_tuned}
\begin{tabular}{l S S S S S S S S S S S}
\multirow{2}{*}{Model} &
\multicolumn{5}{c}{Base} &
\multicolumn{5}{c}{Tuned} \\
\cmidrule(lr){2-6}\cmidrule(lr){7-11}
& {Soybean} & {Cotton} & {Rice} & {Corn} & {Total}
& {Soybean} & {Cotton} & {Rice} & {Corn} & {Total} & {$\Delta$}\\
\midrule

\addlinespace[2pt]
\multicolumn{11}{l}{\textbf{Gemma3}} \\
Gemma3-4B
& 56.56 & 55.47 & 43.31 & 52.76 & 51.98
& 68.85 & 63.28 & 56.69 & 66.14
& \multicolumn{1}{c}{63.69$^{**}$} 
& 11.71 \\

Gemma3-12B
& 69.67 & 62.50 & 56.69 & 64.57 & 63.29
& 77.87 & 72.66 & 62.99 & 70.87
& \multicolumn{1}{c}{71.03$^{**}$} 
& 7.74 \\

Gemma3-27B
& 71.31 & 67.19 & 58.27 & 70.87 & 66.87
& 80.33 & 75.78 & 66.14 & 71.65
& \multicolumn{1}{c}{73.41$^{**}$} 
& 6.55 \\

\addlinespace[4pt]
\multicolumn{11}{l}{\textbf{Qwen3-VL-Instruct}} \\
Qwen3-VL-2B
& 69.13 & 64.06 & 53.81 & 62.99 & 62.43
& 76.23 & 76.56 & 63.78 & 70.08
& \multicolumn{1}{c}{71.63$^{**}$} 
& 9.19\\

Qwen3-VL-4B
& 72.40 & 71.88 & 67.19 & 62.99 & 68.58
& 72.95 & 76.82 & 69.82 & 64.83 & 71.10 & 2.51\\

Qwen3-VL-8B
& 72.13 & 75.78 & 69.29 & 65.35 & 70.63
& 79.51 & 78.12 & 74.02 & 74.02
& \multicolumn{1}{c}{76.39$^{*}$} 
& 5.75\\

Qwen3-VL-30B
& 73.50 & 73.96 & 69.29 & 70.60 & 71.83
& 77.87 & 75.00 & 74.02 & 73.23 & 75.00 & 3.17\\

\addlinespace[4pt]
\multicolumn{11}{l}{\textbf{Other VLMs}} \\
InternVL3.5-2B
& 70.49 & 63.28 & 48.03 & 65.35 & 61.71
& 77.60 & 71.61 & 58.01 & 68.77
& \multicolumn{1}{c}{68.92$^{**}$} 
& 7.21 \\

Granite-Vision
& 53.55 & 49.48 & 44.36 & 48.82 & 49.01
& 63.93 & 66.41 & 52.76 & 56.69
& \multicolumn{1}{c}{59.92$^{**}$} 
& 10.91 \\

LLaVA-8B
& 49.18 & 44.53 & 53.54 & 51.97 & 49.80
& 68.03 & 61.98 & 55.38 & 58.27
& \multicolumn{1}{c}{60.85$^{**}$} 
& 11.04 \\

LLaVA-34B
& 57.38 & 61.20 & 53.02 & 57.74 & 57.34
& 80.33 & 70.31 & 62.99 & 74.02
& \multicolumn{1}{c}{71.83$^{**}$} 
& 14.48 \\

\end{tabular}
\par\vspace{2pt}
\footnotesize
\begin{minipage}{\textwidth}
\raggedright
\text{Note:}  Base: original released checkpoint without any additional training; Tuned: Base model further fine-tuned on our \emph{silver} split. 
Stars denote statistically significant improvements of the Tuned model
over the corresponding Base model after Holm correction:
$^{*}p_{\mathrm{Holm}}<0.05$ and $^{**}p_{\mathrm{Holm}}<0.01$.
\end{minipage}
\end{table*}

\section{Experimental Setup, Results, and Error Analysis}

This section evaluates evidence-grounded multimodal reasoning capabilities on our proposed benchmark, PlantXpert, and examines how agronomic adaptation changes VLM performance in plant phenotyping tasks.
Our experiments are designed not merely to compare model performance, but to answer four research questions (RQs) that characterize multimodal model reasoning in plant phenotyping:
\textbf{RQ1:} \textit{To what extent does task-specific fine-tuning improve evidence-grounded agronomic reasoning?}
\textbf{RQ2:} \textit{How does model scale interact with domain adaptation to determine benchmark performance?}
\textbf{RQ3:} \textit{How robustly do current VLMs generalize across different crop domains?}
\textbf{RQ4:} \textit{Which agronomic domains and reasoning challenges remain bottlenecks even after tuning?}
To maintain consistent comparison, we performed all model evaluations on the expert-validated Gold split under a unified multiple-choice protocol, with results averaged across multiple repeated runs.
Beyond aggregate performance, we further analyze model behavior through challenge-specific, domain-specific, and qualitative error analysis, allowing us to examine not only whether models improve after adaptation, but also where and why important reasoning failures persist.

\paragraph{Evaluation settings.}
The Silver split is used for training, and is disjoint with the evaluation set (the Gold split) at both question and figure level. 
We evaluate the models exclusively with the Gold split with a total of 504 expert-annotated questions: 122 for soybean, 128 for cotton, and 127 for rice and corn. 
During evaluation, the input consists only of the scientific figure and its corresponding multiple-choice question and answer options. 
Captions, related paragraphs, source-paper text, and structured metadata are not provided at inference time.
To keep in line with previous agricultural benchmark studies~\citep{Shinoda_ICCV_2025, yan_agrieval_2026, Arshad_WACV_2025}, we evaluated model performance using multiple-choice accuracy, which is calculated as the proportion of questions for which the selected answer matches the gold label. 
To ensure deterministic and consistent answer generation across models, we used deterministic decoding \texttt{do\_sample=False}.
All results are averaged over three repeated inference runs.
We apply a simple rule-based parser to extract the answer from model output.
The output is first converted to uppercase, and we then search for standalone occurrences of the letters A, B, C, or D using word-boundary matching. 
This strategy allows outputs such as “A,” “Answer: A,” or “I choose A” to be parsed consistently while avoiding matches to letters appearing inside unrelated words.
If exactly one unique answer choice is detected, that letter is taken as the model prediction, and the prediction is correct if it matches the ground-truth answer from experts.
On the other hand, if no valid answer choice is detected, or if multiple distinct choices appear in the same output, the response is treated as invalid and counted as incorrect.
Furthermore, to assess the statistical significance of the fine-tuning gains, we apply a paired, crop-stratified cluster bootstrap test. 
For each question, we record the average correctness (1 for correct, and 0 for incorrect) across the three runs for both the Base and Tuned settings, and compute their paired difference. 
We then bootstrap with 10,000 resamples to estimate the Base-to-Tuned accuracy difference. Figures are resampled with replacement separately within each crop subset, while all questions derived from a sampled figure are retained together. 
This accounts for within-figure dependence while preserving the benchmark's crop composition despite differences in the number of questions per figure. 
We report two-sided null-centered bootstrap p-values for overall Base-to-Tuned accuracy differences, with the Holm correction applied across the 11 Base-to-Tuned model comparisons.

\subsection{Question–Driven Results}
\paragraph{\textbf{RQ1: To what extent does task-specific fine-tuning improve evidence-grounded agronomic reasoning?}}
We first examine whether task-specific fine-tuning improves overall accuracy and, more importantly, strengthens evidence-grounded agronomic reasoning in plant phenotyping. 
As shown in \autoref{tab:vlm_0shot_base_vs_tuned}, fine-tuning improves performance across all evaluated models, with 9 of 11 models retaining statistically significant Base-to-Tuned improvements after Holm correction. 
The improvement is observed across different model families and parameter scales, indicating that task-specific supervision provides useful signals beyond what can be captured by general-purpose multimodal pretraining.
Some base models, especially models in the Qwen3-VL-Instruct family, already achieve strong zero-shot performance, while several other VLMs struggle in the zero-shot setting with accuracy of around 50\%.
After fine-tuning, all models reach an accuracy of around 60\%, with most exceeding 70\%.
This result suggests that even though modern multimodal models possess substantial transferable visual priors, their broad multimodal competence does not fully substitute for agronomic specialization, and task-specific fine-tuning can effectively adapt these models to agronomic visual reasoning.
In other words, PlantXpert does not test generic visual recognition. 
It also probes task-specific reasoning that remains only partially covered by large-scale pretraining.
However, we also notice that although all models benefit from fine-tuning, the magnitude of improvement is not uniform across model families. 
For example, the LLaVA-Next family shows particularly large gains (e.g., +14.48 percentage points for LLaVA-34B), while models with similar parameter size but from different families improve much more modestly (e.g., +3.17 percentage points for Qwen3-VL-30B). 
Notably, the Qwen3-VL models also have relatively stronger Base performance, suggesting that their smaller marginal gains may partly reflect stronger zero-shot alignment with the task and consequently less remaining headroom for adaptation. 
In contrast, models with weaker initial performance often benefit more substantially from task-specific supervision.
Overall, the answer to RQ1 is clear: our task-specific fine-tuning yields meaningful and substantial gains beyond zero-shot multimodal competence. It remains an effective strategy to improve model performance in plant phenotyping.

\paragraph{\textbf{RQ2: How does model scale interact with domain adaptation in determining benchmark performance?}}
Next, we examine the role of the model scale. 
In general, within the same model collection, larger models tend to outperform smaller ones in the Base setting.
Performance gaps are substantial when the parameter size is relatively small, but diminish considerably as the number of parameters continues to increase.
For example, Gemma3-12B substantially outperforms Gemma3-4B (63.29\% vs. 51.98\%) in the Base setting, but has similar performance with Gemma3-27B (66.87\%).
These results indicate that model capacity is important for establishing strong zero-shot multimodal competence, particularly at smaller scales, but its marginal benefit decreases once sufficient capacity is reached.
The relationship between parameter count and performance becomes noticeably weaker in the Tuned setting.
Qwen3-VL-2B and Qwen3-VL-4B achieve similar accuracy (71.63\% vs. 71.10\%), as do Qwen3-VL-8B and Qwen3-VL-30B (76.39\% vs.75.00\%).
This suggests that once models are adapted to the benchmark, simply adding parameters does not automatically translate into stronger agronomic reasoning.
In other words, the remaining benchmark errors are not caused by insufficient model size alone. 
This observation is important for interpreting the benchmark difficulty. 
If performance continued to improve sharply with scale after training, one might conclude that PlantXpert primarily measures raw model capacity. 
Instead, the diminishing returns from scaling indicate that the benchmark captures a more complicated transition from broad multimodal competence to task-grounded scientific reasoning. 
In fact, fine-tuned smaller models can approach or even match substantially larger Base models. For instance, fine-tuned Gemma3-4B reaches 63.69\%, comparable to the 63.29\% Base performance of Gemma3-12B. 
This indicates that task-specific supervision can compensate for a considerable portion of the performance advantage associated with greater model scale.
Therefore, the answer to RQ2 is that scaling primarily strengthens general multimodal competence before adaptation, but exhibits diminishing returns at larger sizes. 
After fine-tuning, benchmark performance depends less consistently on parameter count, suggesting that effective domain adaptation can be at least as important as additional model capacity for evidence-grounded agronomic reasoning.

\paragraph{\textbf{RQ3: How robustly do current VLMs generalize across different crop domains?}}
RQ3 concerns the cross-crop generalization among the four plants. 
As shown in \autoref{tab:vlm_0shot_base_vs_tuned}, model performance is not uniform across crops in both Base and Tuned setting. 
Soybean and cotton-related questions generally yield the strongest results, while rice tends to be the more challenging crop. 
Although fine-tuning substantially improves performance across the crop subsets, it therefore does not fully eliminate the performance gap among them.
After fine-tuning, soybean becomes the subset with the highest accuracy for most models.
Rice, in contrast, remains the crop that most models struggle with even after adaptation.
This persistent asymmetry suggests that cross-crop generalization remains uneven. 
Benchmark difficulty appears to depend not only on low-level image characteristics, but also on the biological structure and agronomic context underlying each crop domain. 
In other words, transferring multimodal competence across crops involves more than recognizing visually similar objects or patterns.
It also requires sensitivity to crop-specific phenotypes, disease manifestations, and scientific frameworks.
At the same time, crop-level differences should be interpreted with caution. 
The four subsets differ with respect to the composition of the questions, the visual distinguishability of the figures, or the complexity of the underlying plant-science context.  
As a result, observed performance gaps cannot be attributed solely to intrinsic differences between crop species.
In particular, the persistent difficulty of the rice subset indicates that strong aggregate performance does not necessarily imply equally robust reasoning across crop domains.
In short, the answer to RQ3 is that current VLMs generalize across crops only partially.
Task-specific fine-tuning improves performance broadly, but meaningful crop-dependent differences remain, which calls for future benchmark expansion and more controlled evaluation.

\begin{table*}[t]
\centering
\small
\setlength{\tabcolsep}{5pt}
\sisetup{
  table-number-alignment = center,
  table-format = 2.2
}

\caption{Accuracy (\%) by challenge type (mean over 3 runs) for the Base and Tuned models.
Avg. $\Delta$ denotes the macro-average of the Tuned-minus-Base accuracy differences across the four challenge types, with each challenge weighted equally.}
\label{tab:challenge_breakdown_0shot_base_vs_tuned}
\begin{tabular}{l S S S S S S S S S}
\multirow{2}{*}{Model} &
\multicolumn{4}{c}{Base} &
\multicolumn{4}{c}{Tuned} \\
\cmidrule(lr){2-5}\cmidrule(lr){6-9}
& {Others} & {Quant.} & {Reasoning} & {Visual} &
  {Others} & {Quant.} & {Reasoning} & {Visual} & {Avg.$\Delta$}\\
\midrule

\addlinespace[2pt]
\multicolumn{9}{l}{\textbf{Gemma3}} \\
Gemma3-4B
& 45.65 & 44.83 & 54.60 & 54.33
& 63.04 & 57.47 & 65.64 & 64.90 
& 12.91\\

Gemma3-12B
& 71.74 & 40.23 & 68.10 & 67.31
& 76.09 & 62.07 & 73.01 & 72.12 
& 8.98\\

Gemma3-27B
& 73.91 & 45.98 & 70.55 & 71.15
& 78.26 & 62.07 & 76.07 & 75.00 
& 7.45\\

\addlinespace[4pt]
\multicolumn{9}{l}{\textbf{Qwen3-VL-Instruct}} \\
Qwen3-VL-2B
& 75.36 & 45.21 & 59.92 & 68.75
& 76.09 & 59.77 & 73.62 & 74.04 
& 8.57\\

Qwen3-VL-4B
& 86.96 & 49.43 & 71.57 & 70.19
& 86.96 & 55.17 & 74.85 & 71.31 
& 2.54\\

Qwen3-VL-8B
& 78.26 & 55.17 & 73.62 & 73.08
& 76.09 & 64.37 & 78.53 & 79.81 
& 4.67\\

Qwen3-VL-30B
& 78.26 & 55.17 & 75.05 & 74.84
& 82.61 & 60.92 & 76.07 & 78.37 
& 3.66\\

\addlinespace[4pt]
\multicolumn{9}{l}{\textbf{Other VLMs}} \\
InternVL3.5-2B
& 67.39 & 41.38 & 64.42 & 66.83
& 79.71 & 55.56 & 68.10 & 72.76 
& 9.03\\

Granite-Vision
& 45.65 & 33.33 & 54.81 & 51.76
& 63.04 & 45.98 & 63.19 & 62.50 
& 12.29\\

LLaVA-8B
& 50.00 & 19.54 & 56.44 & 57.21
& 65.94 & 42.53 & 62.17 & 66.35 
& 13.45\\

LLaVA-34B
& 60.87 & 35.63 & 59.92 & 63.62
& 82.61 & 55.17 & 70.55 & 77.40 
& 16.42\\

\end{tabular}%
\par\vspace{2pt}
\footnotesize
\begin{minipage}{\textwidth}
\raggedright
\text{Note:} Base: the original released checkpoint without additional training; 
Tuned: the Base model further fine-tuned on the \emph{silver} split. 
Challenge types include Visual Perception \& Expertise, Reasoning \& Logic, 
Quantitative \& Statistical Estimation, and Others.
\end{minipage}
\end{table*}

\paragraph{\textbf{RQ4: Which agronomic domains and reasoning challenges remain bottlenecks even after tuning?}}
Our final question asks which reasoning challenges and agronomic domains remain difficult even after fine-tuning. 
As shown in \autoref{tab:challenge_breakdown_0shot_base_vs_tuned} and \autoref{tab:domain_breakdown_0shot_base_vs_tuned}, task-specific fine-tuning generally improves performance across challenge types and agronomic domains, although the magnitude of improvement varies considerably across models. 
For example, LLaVA-34B shows the largest macro-average improvement across challenge types (+16.42 percentage points) and one of the largest improvements across domains (+12.99 percentage points), whereas Qwen3-VL-4B (only +2.54 percentage points across challenge types and +2.26 across domains) and Qwen3-VL-30B exhibit substantially smaller average gains. 
Despite these broad improvements, several persistent bottlenecks remain, not in basic visual recognition, but in more structured forms of agronomic inference.
We first consider challenge types. 
\autoref{tab:challenge_breakdown_0shot_base_vs_tuned} shows that quantitative reasoning is consistently the most difficult category across models, with Base performance in most cases falling below 50\% (only 19.54\% for LLaVA-8B). 
Although fine-tuning substantially improves performance for this category, the gap relative to other challenge types remains substantial.
This indicates persistent difficulty with quantitative and statistical estimation tasks. 
By contrast, Reasoning \& Logic and Visual Perception \& Expertise generally achieve considerably higher accuracy, with many models exceeding 70\% after training. 
Thus, the remaining challenge is not simply recognizing relevant visual content, but transforming that evidence into precise quantitative conclusions.
We also work to identify patterns across the agronomic domains. 
\autoref{tab:domain_breakdown_0shot_base_vs_tuned} shows that Remote Sensing \& Imaging appears to be one of the most challenging domains for models under both settings, with accuracy remaining around 50\% in several cases even after training. 
Meanwhile, it is noteworthy that fine-tuning generally narrows the differences among the other agronomic domains. 
Crop Science \& Agronomy, Plant Biology \& Physiology, and Plant Health \& Pathology subsets all show strong post-tuning performance in many cases.
These improvements suggest that task-specific supervision helps models connect visual evidence with biological and agronomic knowledge rather than relying only on surface-level visual cues.
In other words, our task-specific fine-tuning is effective in improving the models' capability to move beyond surface-level visual recognition and perform evidence-grounded phenotypic reasoning.
On the other hand, remote sensing and imaging questions can require spatial pattern interpretation, spectral and sensor-derived representations, and segmentation and detection outputs. 
Such information is less directly represented by ordinary image-based pretraining, which may be a factor behind the low accuracy in this domain.
In general, the answer to RQ4 is that the main remaining bottlenecks lie in quantitative reasoning and remote sensing \& imaging interpretation.
Fine-tuning improves performance across challenge types and domains, but only partially eliminates performance disparities. 
Still, task-specific training remains an effective method to achieve deeper agronomic reasoning capabilities on current multimodal models.

Together, the results present a consistent picture of the current VLM performance in plant phenotyping. 
Task-specific fine-tuning yields meaningful gains beyond zero-shot performance, while increasing model scale alone yields diminishing returns.
The cross-crop generalization remains uneven, and the most persistent weaknesses lie in quantitative reasoning as well as specialized scientific image interpretation. 
Overall, the broad improvements following fine-tuning demonstrate the potential of task-specific supervision to strengthen evidence-grounded agronomic reasoning capabilities that are only partially captured by general-purpose multimodal pretraining.

\begin{table*}[t]
\centering
\small
\setlength{\tabcolsep}{4pt}
\sisetup{
  table-number-alignment = center,
  table-format = 2.2
}

\caption{Accuracy (\%) by domain (mean over 3 repeated inference runs) for the Base and Tuned models.
Avg. $\Delta$ denotes the macro-average of the Tuned-minus-Base accuracy differences across the six domains, with each domain
weighted equally.}
\label{tab:domain_breakdown_0shot_base_vs_tuned}
\resizebox{0.95\textwidth}{!}{%
\begin{tabular}{l S S S S S S S S S S S S S}

\multirow{2}{*}{Model} &
\multicolumn{6}{c}{Base} &
\multicolumn{6}{c}{Tuned} \\
\cmidrule(lr){2-7}\cmidrule(lr){8-13}
& {CropSci.} & {Entom.} & {Others} & {Bio.} & {Health} & {Remote} &
  {CropSci.} & {Entom.} & {Others} & {Bio.} & {Health} & {Remote} & {Avg. $\Delta$}\\
\midrule

\addlinespace[2pt]
\multicolumn{13}{l}{\textbf{Gemma3}} \\

Gemma3-4B
& 51.95 & 48.78 & 42.86 & 56.34 & 53.37 & 50.77
& 66.23 & 51.22 & 61.90 & 69.01 & 66.83 & 53.85 & 10.83\\

Gemma3-12B
& 61.04 & 53.66 & 73.81 & 70.42 & 65.38 & 50.77
& 74.03 & 63.41 & 69.05 & 67.61 & 75.00 & 64.62 & 6.44\\

Gemma3-27B
& 64.94 & 63.41 & 76.19 & 66.20 & 70.19 & 55.38
& 72.73 & 65.85 & 76.19 & 74.65 & 79.81 & 55.38 & 4.72\\

\addlinespace[4pt]
\multicolumn{13}{l}{\textbf{Qwen3-VL-Instruct}} \\

Qwen3-VL-2B
& 58.44 & 61.79 & 58.73 & 54.93 & 69.55 & 55.38
& 70.13 & 63.41 & 73.81 & 70.42 & 75.48 & 66.15 & 10.10\\

Qwen3-VL-4B
& 70.13 & 70.73 & 73.81 & 62.44 & 71.79 & 58.46
& 71.00 & 74.80 & 69.05 & 73.71 & 73.88 & 58.46 & 2.26\\

Qwen3-VL-8B
& 64.94 & 63.41 & 71.43 & 69.01 & 75.00 & 69.23
& 75.32 & 73.17 & 73.81 & 77.46 & 79.33 & 70.77 & 6.14\\

Qwen3-VL-30B
& 70.56 & 68.29 & 70.63 & 67.61 & 76.92 & 64.62
& 75.32 & 68.29 & 76.19 & 71.83 & 79.81 & 66.15 & +3.16\\

\addlinespace[4pt]
\multicolumn{13}{l}{\textbf{Other VLMs}} \\

InternVL3.5-2B
& 50.65 & 53.66 & 73.81 & 60.56 & 69.71 & 47.69
& 62.77 & 57.72 & 78.57 & 71.83 & 74.68 & 55.38 & 7.48\\

Granite-Vision
& 43.29 & 47.97 & 38.89 & 52.11 & 54.81 & 41.03
& 46.75 & 60.98 & 61.90 & 64.79 & 65.38 & 50.77 & 12.08\\

LLaVA-8B
& 50.65 & 48.78 & 50.00 & 45.07 & 55.29 & 36.92
& 52.38 & 58.54 & 62.70 & 60.56 & 68.27 & 47.69 & +10.57\\

LLaVA-34B
& 46.32 & 68.29 & 54.76 & 57.28 & 61.54 & 51.79
& 64.94 & 73.17 & 71.43 & 74.65 & 78.37 & 55.38 & 12.99\\

\end{tabular}%
}
\par\vspace{2pt}
\footnotesize
\begin{minipage}{\textwidth}
\raggedright
\text{Note:} Base: the original released checkpoint without any additional training; Tuned: the Base model further fine-tuned on our \emph{silver} split. Domains included: Crop Science \& Agronomy, Entomology \& Weed Science, Others, Plant Biology \& Physiology, Plant Health \& Pathology, and Remote Sensing \& Imaging.
\end{minipage}
\end{table*}

\subsection{Error Analysis and Case Studies}

\begin{figure*}[]
\centering
\includegraphics[width=0.95 \textwidth]{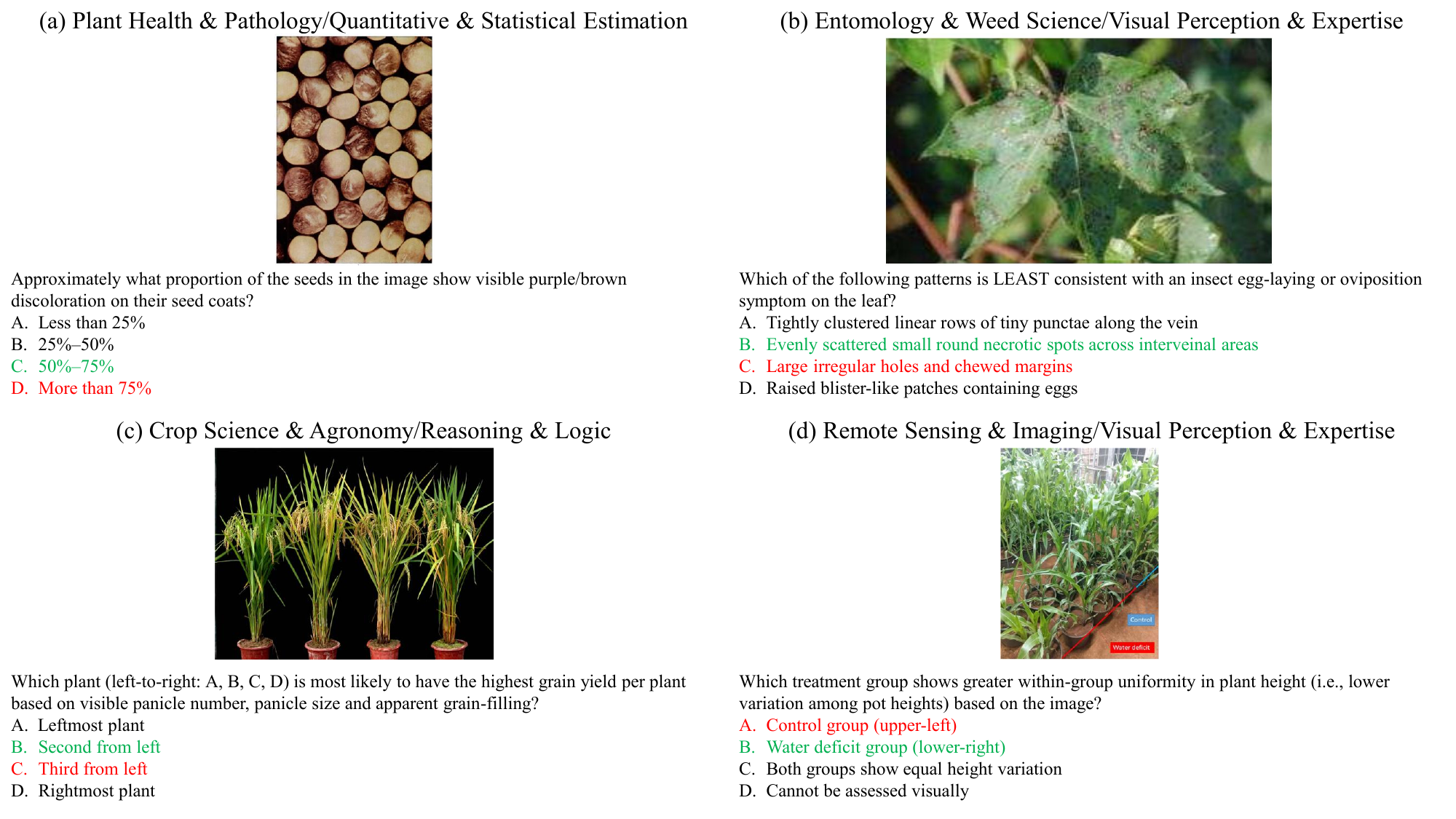} 
\caption{Representative persistent error cases from the task-specific Tuned models.}
\label{fig:wrong}
\end{figure*}

To better understand the limitations of current VLMs in this benchmark, we further examine common types of incorrect predictions from our Tuned models in the Gold split. 
We focus on persistent errors for which the Tuned model produced an incorrect prediction in all three repeated inference runs.
We group the wrong examples by their corresponding challenge types and agronomic domains, and select representative cases to illustrate typical failure modes in \autoref{fig:wrong}, where green denotes the expert-validated answer and red denotes the incorrect model prediction.
One major source of errors lies in quantitative reasoning questions. 
In the example provided in \autoref{fig:wrong} (a), the task requires more than detecting whether discoloration is present. 
Instead, the model must identify affected seeds across the image, implicitly count or estimate their frequency, and map that estimate to a predefined percentage range.
Such errors are consistent with our previous results, where quantitative reasoning questions remain the lowest-performing category for every model after fine-tuning.
This suggests that VLMs can often recognize relevant phenotypic symptoms while still struggling to convert visual observations into reliable numerical judgments. 
We also investigate errors across other domain and challenge categories.
As shown in \autoref{fig:wrong} (b), unlike ordinary visual perception tasks, Entomology \& Weed Science questions often require fine-grained discrimination between visually plausible symptoms and knowledge of their likely biological causes.
The failure therefore illustrates that strong visual recognition does not necessarily compensate for insufficient domain expertise.
In real-life plant phenotyping tasks, models must not only distinguish between similar plant damage patterns but also associate them with the correct entomological interpretation.

A different type of failure appears in Crop Science \& Agronomy reasoning, as illustrated in \autoref{fig:wrong}(c).
Here there is no single visual feature in the image that can directly determine the answer. 
Instead, the model must integrate several phenotypic cues and use them as proxies for an agronomically meaningful outcome, in this case grain yield. 
This case illustrates the difficulty of moving from local visual recognition to higher-level scientific inference. 
A model may correctly perceive individual plant characteristics yet still fail to combine them appropriately when making a biologically grounded judgment.
Finally, \autoref{fig:wrong}(d) illustrates the persistent difficulty in the Remote Sensing \& Imaging domain, which emerges as one of the most challenging domains in our previous results.
Although categorized as Visual Perception \& Expertise, this question implicitly requires a form of spatial reasoning.
The model must identify the relevant groups, compare the distribution of plant heights within each group, and determine which exhibits lower variability. 
It helps explain why specialized imaging questions can remain difficult even when general visual-perception performance is relatively strong.
In general, these observations indicate that current VLMs after task-specific fine-tuning have strong capabilities in identifying meaningful phenotypic evidence.
However, they can still fail at different stages of the evidence-grounded reasoning process before producing the correct conclusion.
Therefore, more task-specific training will be important to improve the capability of VLMs for agricultural applications.
Future work will require training strategies that go beyond additional examples of object recognition and more explicitly support quantitative reasoning, fine-grained domain expertise, and multi-step evidence-grounded inference.

\section{Robustness and Ablation Studies}

Beyond the main benchmark results, we conduct a series of additional analyses to examine whether the observed performance trends remain stable under stricter evaluation conditions and whether model predictions are genuinely supported by the visual input. 
Specifically, we evaluate source-paper out-of-distribution generalization, training-seed robustness, sensitivity to visual and answer-choice perturbations, expert-assessed rationale grounding, and dependence on the synthetic data generator on a representative subset of models. 
Together, these experiments are designed to test the robustness of our conclusions and to identify potential sources of bias or shortcut behavior that may not be visible from aggregate accuracy alone.

\subsection{Source-Paper Out-Of-Distribution Generalization}

Although the original Gold split is disjoint from the training data at the question and figure levels, questions derived from different figures of the same source paper may still appear across the training and evaluation sets. 
This leaves a meaningful residual dependency because figures from one paper are likely to have the same crop, environment, or plant condition.
To examine whether such source-level overlap inflates performance, we construct a stricter \textit{Gold-OOD} subset by removing every Gold sample whose source paper appears in the full four-crop training set. 
Source papers are matched using the combination of crop type and paper ID to account for crop-specific paper indexing. 
This filtering reduces the original 504-question \textit{Gold} set to 91 questions from previously unseen source papers.
We evaluate three representative fine-tuned VLMs on this stricter subset and compare their performance with that on the original Gold set. 
As shown in Table~\ref{tab:paper_ood}, performance decreases only moderately under the stricter source-paper separation. 
Qwen3-VL-8B decreases from 76.39\% to 71.43\% ($-4.96$ percentage points), while LLaVA-Next-8B decreases from 60.85\% to 56.04\% ($-4.81$ points). Gemma-3-12B remains nearly unchanged, decreasing from 71.03\% to 70.33\% ($-0.70$ points). 
Across the three representative models, the average decrease is 3.49 percentage points. 
These results suggest that the observed performance can not be solely attributed to source-paper level overlap and that the models retain substantial predictive ability on questions derived from entirely unseen source papers.

\begin{table}[t]
\centering
\caption{Accuracy comparison between the original figure-disjoint Gold set (\textit{Gold}) and the stricter source-paper-disjoint subset (\textit{Gold-OOD}). $\Delta$ denotes the change from Gold-Figure to Gold-Paper in percentage points.}
\label{tab:paper_ood}
\begin{tabular}{lccc}
\toprule
Model & Gold & Gold-OOD & $\Delta$ \\
\midrule
Qwen3-VL-8B     & 76.39 & 71.43 & -4.96 \\
LLaVA-Next-8B   & 60.85 & 56.04 & -4.81 \\
Gemma-3-12B     & 71.03 & 70.33 & -0.70 \\
\bottomrule
\end{tabular}
\end{table}

\subsection{Training-Seed Robustness}

Our main fine-tuning results show consistent improvements over the Base models.
However, they are obtained from a single training run per model and do not provide a meaningful estimate of potential variability arising from the fine-tuning process.
To examine whether the observed fine-tuning gains depend on are actually result of a particular optimization trajectory, we independently fine-tune the three models from the Qwen3-VL family (Qwen3-VL-2B, Qwen3-VL-8B, and Qwen3-VL-30B) using three independent random training seeds (42, 43, and 44). 
These three models are chosen as a representative subset because of their small, medium, and large model scales, respectively. 
All training data, hyperparameters, and evaluation settings are held fixed across seeds.
As shown in Table~\ref{tab:seed_robustness}, the fine-tuned models
consistently outperform their corresponding Base models across all nine independent training runs. 
Qwen3-VL-2B achieves $71.30 \pm 1.62\%$ accuracy, compared with 62.43\% for the Base model, corresponding to an average improvement of 8.87 percentage points.
Qwen3-VL-8B achieves $76.19 \pm 0.40\%$, improving over its 70.63\%
Base accuracy by 5.56 points, while Qwen3-VL-30B achieves
$75.26 \pm 1.09\%$, compared with 71.83\% for the Base model, for an
average gain of 3.43 points. 
The relatively small standard deviations, together with consistent improvements over the Base models for every training seed, indicate that the observed gains are robust to optimization randomness rather than being attributable to a particular training
trajectory.

\begin{table}[t]
\centering
\caption{Training-seed robustness on the Gold evaluation set. 
Each model is independently fine-tuned using three random seeds. 
Accuracy is reported in percent, and $\Delta$ denotes the difference between the mean
fine-tuned accuracy and the corresponding Base accuracy.}

\label{tab:seed_robustness}
\begin{tabular}{lcccccc}
\toprule
Model & Base & Seed 1 & Seed 2 & Seed 3 & Mean $\pm$ SD & $\Delta$ \\
\midrule
Qwen3-VL-2B  & 62.43 & 72.42 & 69.44 & 72.02 & 71.30 $\pm$ 1.62 & +8.87 \\
Qwen3-VL-8B  & 70.63 & 75.79 & 76.19 & 76.59 & 76.19 $\pm$ 0.40 & +5.56 \\
Qwen3-VL-30B & 71.83 & 76.39 & 74.21 & 75.20 & 75.26 $\pm$ 1.09 & +3.43 \\
\bottomrule
\end{tabular}
\end{table}

\subsection{Visual Grounding and Answer-Bias Ablations}
\label{sec:visual_grounding}

Our benchmark is explicitly designed around image-based reasoning, but we can not rule out the possibility of a model answering correctly using elimination, textual priors, option structure, or partial visual recognition without demonstrating any intended agronomic reasoning.
To examine this possibility, we conduct three controlled ablations on the full 504-question Gold set using Qwen3-VL-8B, LLaVA-Next-8B, and Gemma-3-12B. 
We evaluate each tuned model with three diagnostic subsets:
(1) image-masked, in which we remove the image entirely, 
(2) image-shuffled, in which we replace the correct figure with an
unrelated Gold-set figure using a fixed figure-level permutation, and
(3) answer-choice-shuffled, in which we random reorder the four answer choices while remapping the correct answer accordingly. 
As shown in Table~\ref{tab:visual_ablation}, Qwen3-VL-8B exhibits substantial performance decrease when visual information is removed or mismatched, falling from 76.39\% to 59.13\% without an image and to 56.75\% with a shuffled image. 
Gemma-3-12B shows a similar pattern, with its accuracy decreasing from 71.03\% to 57.94\% and 53.77\%, respectively. 
In contrast, LLaVA-Next-8B is relatively less sensitive to the visual perturbations, with accuracy decreasing from 60.85\%
to 56.94\% without the image and 57.34\% with a shuffled image. This suggests that VLMs rely substantially on the correspondence between the question and its visual evidence. 
LLaVA-Next-8B shows a smaller absolute decrease under visual perturbations, but its original accuracy is also significantly lower than that of Qwen3-VL-8B and Gemma-3-12B. 
Therefore, the smaller gap is more likely to reflect a lower initial level of successful visual reasoning rather than stronger robustness to image removal or mismatch.
On the other hand, option shuffling has less negative impact for all three models, with a decrease of 4.76 percentage points for Qwen3-VL-8B, only 2.32 points for LLaVA-Next-8B,
and 6.35 points for Gemma-3-12B. 
Averaged across the three models, no-image and image-shuffle perturbations reduce accuracy by 11.42 and 13.47 points, respectively, compared with a 4.48-point only reduction under
option shuffling. 
Overall, these results indicate that visual evidence
contributes substantially to model performance, and that the VLMs demonstrate their intended agronomic reasoning.

\begin{table}[t]
\centering
\caption{Visual-grounding and answer-bias ablations on the Gold set.
Values are accuracy (\%), with changes relative to the original evaluation
shown in parentheses.}
\label{tab:visual_ablation}
\begin{tabular}{lcccc}
\toprule
Model & Original (Tuned) & Image Mask & Image Shuffle & Option Shuffle \\
\midrule
Qwen3-VL-8B
& 76.39
& 59.13 (-17.26)
& 56.75 (-19.64)
& 71.63 (-4.76) \\

LLaVA-Next-8B
& 60.85
& 56.94 (-3.91)
& 57.34 (-3.51)
& 58.53 (-2.32) \\

Gemma-3-12B
& 71.03
& 57.94 (-13.09)
& 53.77 (-17.26)
& 64.68 (-6.35) \\
\bottomrule
\end{tabular}
\end{table}

\subsection{Expert Evaluation of Rationale Grounding}

To further validate the evidence-grounded reasoning capabilities of VLMs, we randomly sample 50 questions from the 504-question Gold set and generate answer rationales using the same set of representative tuned models in \autoref{sec:visual_grounding} (Qwen3-VL-8B, LLaVA-Next-8B, and Gemma-3-12B), obtaining 150 rationales in total.
We deploy the evaluation through Label Studio ~\citep{Label} and invite two domain experts to independently rate each rationale on a five-point grounding scale. 
Here a score of 1 denotes no meaningful or unsupported rationale, 2 denotes largely incorrect or generic visual evidence, 3 denotes partial grounding, 4 denotes mostly grounded reasoning, and 5 denotes fully and specifically grounded reasoning.
Annotators are provided with the original figure, question and options, model prediction, and generated rationale. 
Rationale grounding is evaluated separately from multiple-choice correctness, allowing us to distinguish correct predictions
that are visually justified from those based on weak or unsupported explanations.
For analysis, responses receiving a score of 1 from both experts are treated as no-rationale cases.
These responses remain part of the overall evaluation but are separated from the calculation of grounding quality, since averaging them together with substantive rationales would conflate failure to generate an explanation with the quality of an explanation that is actually provided.
For the remaining rationale-bearing responses, we average the two expert scores and consider a rationale grounded when its mean grounding score is at least 4.
We additionally report the proportion of responses that are both multiple-choice correct and grounded across the complete 50-question sample.
The two experts achieve 88.0\% exact agreement across the 150 model responses, with a linearly weighted Cohen's $\kappa$ of 0.932, indicating strong agreement on the ordinal grounding scale.
As shown in Table~\ref{tab:rationale_grounding}, all three models achieve the same multiple-choice accuracy of 76.00\% on the sampled 50 questions, but differ substantially in their ability to consistently provide grounded rationales.
Qwen3-VL-8B does not provide a meaningful rationale for 78.00\% of the sampled questions even though it is explicitly prompted to do so.
However, among the 11 responses for which a rationale is provided, its mean grounding score is 4.55 and 90.91\% are rated as grounded.
LLaVA-Next-8B shows a lower no-rationale rate of 44.00\%, while its 28 rationale-bearing responses achieve a mean grounding score of 4.52 and a grounded rate of 85.71\%.
Gemma-3-12B provides rationales most consistently, with a no-rationale rate of only 12.00\%.
Among its 44 rationale-bearing responses, the mean grounding score is 4.38 and 72.73\% are considered grounded.
The difference becomes particularly clear when answer correctness and rationale grounding are considered jointly.
Gemma-3-12B produces responses that are both correct and grounded for 64.00\% of the complete sample, compared with 44.00\% for LLaVA-Next-8B and 18.00\% for Qwen3-VL-8B.
These results demonstrate that multiple-choice accuracy alone does not fully characterize evidence-grounded reasoning.
In particular, Qwen3-VL-8B produces strongly grounded explanations when a rationale is present but frequently provides only an answer, whereas Gemma-3-12B more consistently combines correct predictions with meaningful visually grounded explanations.
Overall, although rationale generation remains sensitive to prompting and some models frequently fail to provide an explanation, we can still directly assess whether model predictions are supported by meaningful visual evidence by separating these no-rationale cases from the grounding quality of substantive responses.
These results provide additional validation for the central term of evidence-grounded reasoning beyond multiple-choice accuracy.

\begin{table}[t]
\centering
\caption{Expert evaluation of rationale grounding on 50 randomly sampled Gold questions. No Rationale denotes responses assigned a grounding score of 1 by both experts. Mean Grounding and Grounded are calculated only among rationale-bearing responses, where Grounded denotes a mean expert score of at least 4. Correct + Grounded is calculated over the complete 50-question sample.}
\label{tab:rationale_grounding}
\begin{tabular}{lccccc}
\toprule
\textbf{Model} &
\textbf{MCQ Acc.} &
\textbf{No Rationale} &
\textbf{Mean Grounding} &
\textbf{Grounded} &
\textbf{Correct + Grounded} \\
\midrule
Qwen3-VL-8B
& 76.00\%
& 78.00\%
& 4.55
& 90.91\%
& 18.00\% \\

LLaVA-Next-8B
& 76.00\%
& 44.00\%
& 4.52
& 85.71\%
& 44.00\% \\

Gemma-3-12B
& 76.00\%
& 12.00\%
& 4.38
& 72.73\%
& 64.00\% \\
\bottomrule
\end{tabular}
\end{table}

\subsection{Synthetic-Generator Sensitivity}

In the synthetic question-generation pipeline, we use a fixed prompt and structured output format.
However, relying on a single generator can introduce generator-specific linguistic patterns, distractor styles, or other distributional artifacts. 
To examine this sensitivity, we randomly select 10\% of the source figures (54 in total) and construct two matched training subsets using the original
GPT-5-mini \citep{singh_openai_2025} and Claude Sonnet 5.5 \citep{anthropic_claude_2026}, following the same pipeline. 
To rule out direct evaluation leakage, we exclude any figure that appears in the Gold set during subset construction.
We fine-tune Qwen3-VL-8B separately on the two matched subsets using identical training settings and evaluate both models on the same 504-question human-validated Gold set.
As shown in Table~\ref{tab:generator_sensitivity}, the choice of synthetic generator has a substantial effect on downstream performance.
Starting from the same Qwen3-VL-8B Base model with 70.63\% accuracy, fine-tuning on the GPT-5-mini-generated subset improves accuracy to 76.98\%, corresponding to a gain of 6.35 percentage points.
In contrast, fine-tuning on the matched Claude Sonnet 5.5-generated subset results in 69.84\% accuracy, 0.79 percentage points below the Base model.
This produces a 7.14-point performance difference between the two generator-specific models.
These results suggest that downstream performance can be affected by the synthetic data generator.
However, it is noteworthy that Gold questions originate from the GPT-based generation pipeline, and the residual stylistic or structural alignment may favor GPT-generated training data.
We therefore interpret the results as evidence of generator sensitivity rather than attribute the performance improvement to generator-specific artifacts.

\begin{table}[t]
\centering
\caption{Synthetic-generator sensitivity of Qwen3-VL-8B. Both fine-tuning
conditions use matched source figures, equal training-set size, and identical
training settings.}
\label{tab:generator_sensitivity}
\begin{tabular}{lcc}
\toprule
\textbf{Training Data} & \textbf{Gold Accuracy (\%)} &
\textbf{$\Delta$ vs. Base (pp)} \\
\midrule
Base & 70.63 & -- \\
GPT-5-mini & 76.98 & +6.35 \\
Claude Sonnet 5.5 & 69.84 & -0.79 \\
\bottomrule
\end{tabular}
\end{table}

\subsection{Quantitative Difficulty Analysis}

In PlantXpert, we provide valuable metadata which include qualitative descriptions of question difficulty.
However, these metadata do not provide a directly measurable quantitative characterization of benchmark complexity given the considerable variation among scientific figures.
We therefore complement the existing annotations with
three difficulty descriptors that capture both textual and visual
properties of the Gold set, question length, figure resolution, and image entropy.
Question length is measured by the number of words in the question text and serves as a simple proxy for textual complexity. 
Figure resolution is measured using the pixel dimensions of the associated image and summarized in megapixels, providing an objective measure of image
size.
We additionally compute grayscale Shannon entropy for each figure as a proxy for visual complexity, where higher entropy
indicates greater variation in pixel intensities.
To examine whether these descriptors are associated with model performance, we analyze predictions from the same three representative tuned models used in
\autoref{sec:visual_grounding}: Qwen3-VL-8B, LLaVA-Next-8B, and Gemma-3-12B. 
We first divide each descriptor into low, medium, and high tertiles based on the 504 Gold questions and compare model accuracy across these groups. 
We further fit a logistic generalized estimating equation (GEE) model with prediction correctness as the binary outcome, the three standardized
difficulty descriptors as continuous predictors, model identity as a fixed effect, and scientific figure as the clustering unit. 
This accounts for the dependence among multiple questions and model predictions associated with the
same figure.
The three descriptors exhibit only weak pairwise correlations ($|\rho| \leq 0.214$), suggesting that they capture relatively distinct properties of benchmark composition. 
Among them, figure resolution shows the
clearest association with model performance. Accuracy decreases consistently from the low- to high-resolution tertile for all three models: from 77.98\% to 74.40\% for Qwen3-VL-8B, from 65.48\% to 54.76\% for LLaVA-Next-8B, and from 76.79\% to 66.67\% for Gemma-3-12B. 
Averaged across the three models, accuracy decreases from 73.42\% in the low-resolution group to 65.28\% in the high-resolution group.
The GEE analysis further supports this association. 
As summarized in Table~\ref{tab:difficulty_analysis}, a one-standard-deviation increase in figure resolution is associated with less accurate prediction
(OR = 0.752, 95\% CI: 0.633--0.893, $p=0.0012$). 
In contrast, question length (OR = 1.108, $p=0.159$) and image entropy (OR = 1.102, $p=0.261$) do not show statistically significant associations
with correctness. 
These results suggest that simple textual length and pixel-level entropy alone do not strongly explain benchmark difficulty, whereas higher-resolution figures are associated with lower model accuracy.

\begin{table}[t]
\centering
\caption{Quantitative difficulty descriptors and their association with model
correctness on the Gold set. Odds ratios (ORs) are obtained from a logistic
GEE model using standardized predictors, model identity as a fixed effect, and
scientific figure as the clustering unit.}
\label{tab:difficulty_analysis}
\begin{tabular}{lcccc}
\toprule
\textbf{Descriptor} &
\textbf{Mean $\pm$ SD} &
\textbf{OR} &
\textbf{95\% CI} &
\textbf{$p$-value} \\
\midrule
Question length (words)
& $22.12 \pm 6.07$
& 1.108
& 0.961--1.278
& 0.159 \\

Figure resolution (MP)
& $0.915 \pm 0.833$
& 0.752
& 0.633--0.893
& 0.0012 \\

Image entropy
& $6.546 \pm 1.217$
& 1.102
& 0.930--1.306
& 0.261 \\
\bottomrule
\end{tabular}
\end{table}

\section{Related Work}

In recent years, a growing number of image datasets have been developed to support AI applications in plant science, particularly for phenotyping and disease-related tasks. 
Early resources such as PlantVillage \citep{mohanty_using_2016} advanced plant disease recognition, but were collected largely in laboratory-style settings with clean backgrounds and limited field variability, limiting their value for in-field use. 
More recent datasets, including PhenoBench \citep{Weyler_PhenoBench_2024}, PlantDoc \citep{singh_plantdoc_2020}, and PlantSeg \citep{wei_plantseg_2024}, improve realism or annotation quality, yet remain focused mainly on perception-level tasks such as recognition, localization, and segmentation rather than higher-level phenotypic interpretation. 
As VLMs have become increasingly prominent in agricultural AI, several studies have introduced multimodal instruction-tuning resources to adapt these models to agricultural tasks, including Agri-LLaVA \citep{wang_agri-llava_2024}, AgroInstruct \citep{awais_agrogpt_2025}, CDDM \citep{liu_multimodal_2025}, and Agri-342k \citep{yang_agrigpt_2025}. 
However, these resources remain focused mainly on perception-level  agricultural tasks, with limited coverage of broader phenotyping scenarios such as plant physiology, developmental interpretation, and scientific figure understanding.
To enable more systematic evaluation, a related line of research has shifted from dataset construction toward benchmark design for assessing VLMs in plant science and agricultural settings.
AgriBench \citep{zhou_agribench_2025} presents a structured evaluation framework spanning tasks from crop recognition to decision-oriented analysis, whereas AgEval \citep{Arshad_WACV_2025} focuses more specifically on plant stress phenotyping through question-answer evaluation. 
Other benchmarks, such as AgroBench \citep{Shinoda_ICCV_2025} and AgriEval \citep{yan_agrieval_2026}, further expand this direction by incorporating expert validation or broadening the scope of evaluated categories. 
Despite these advances, several limitations remain. 
Existing benchmarks are still oriented primarily toward zero-shot or few-shot assessment, providing limited support for standardized supervised adaptation and thus restricting analysis of how VLMs improve under domain-specific tuning. 
Moreover, their structured annotations are often relatively coarse, which limits fine-grained diagnostic analysis across plant science contexts. 
Benchmark construction is also not always explicitly grounded in scientific visual evidence and its accompanying textual interpretation, making it difficult to determine whether model performance reflects evidence-grounded reasoning or merely surface-level recognition. 
To address these gaps, we introduce a plant-phenotyping benchmark that supports controlled adaptation and expert-validated evaluation, while enabling fine-grained analysis of evidence-grounded reasoning over scientific figures.

\section{Conclusion}

We present PlantXpert, an evidence-grounded benchmark for evaluating multimodal reasoning over scientific plant imagery. 
Constructed with plant imagery extracted from publications, PlantXpert organizes visual evidence from figures together with captions and contextual text into agronomically grounded question-answer pairs spanning diverse domains and reasoning challenges. 
Our experiments show that task-specific fine-tuning consistently improves performance across all 11 evaluated VLMs, with 9 models retaining statistically significant Base-to-Tuned improvements after Holm correction. 
These gains are observed across different model families and parameter scales, demonstrating the value of domain-specific supervision to improve evidence-grounded agronomic reasoning capabilities.
At the same time, the benefits of model scaling diminish at larger sizes, and smaller fine-tuned models can approach or even match substantially larger Base models in accuracy, suggesting that model capacity alone is insufficient to address the remaining reasoning challenges.
Our results further reveal that model performance remains uneven across crop subsets, indicating that strong aggregate performance does not necessarily translate into equally robust reasoning across different crop contexts. 
Fine-tuning reduces many of these disparities, but does not eliminate them. 
Across reasoning challenges, Quantitative \& Statistical Estimation remains the clearest bottleneck, with models continuing to struggle when visual evidence must be translated into counting, proportion estimation, or other numerical judgments. 
At the domain level, Remote Sensing and vision inference also remains challenging for most models, highlighting difficulties in interpreting specialized scientific imagery and spatial or sensor-derived visual representations.
In addition, the error analysis shows that persistent failures extend beyond these categories.
Models still have difficulties when it comes to more complex agronomic reasoning, such as fine-grained symptom discrimination, integration of phenotypic cues and latent inference of agronomic outcomes. 
Taken together, these results suggest that current VLMs already possess substantial transferable multimodal competence, but still fall short when visual evidence must be converted into precise quantitative estimates, specialized agricultural reasoning, and higher-level agronomic interpretation.

At the same time, the current version of PlantXpert has several limitations. 
The dataset covers four of the major crops: soybean, cotton, rice, and corn.
However, the novelty of our scientific figure retrieval approach and benchmark construction pipeline also present significant challenges in maintaining balance across crops, figure types, and image quality. 
In addition, the current benchmark remains modest in scale, consisting of 545 images and 4{,}351 QA pairs, with expert annotation available for 504 questions in the Gold split. 
These differences in subset composition should also be taken into account when interpreting crop- and domain-level performance comparisons.
In future work, we plan to expand PlantXpert to additional crops, improve balance across different crop species and visual formats, further increase the scale of the collected scientific images, and continue collaboration with agronomy experts. 
We also plan to introduce an additional external validation set using independent plant-phenotyping images or scientific figures outside the current benchmark collection, including publications released after the original collection cutoff, to evaluate the generalization of Base and Tuned models to previously unseen plant-phenotyping evidence without additional adaptation.
A list of candidate publications for constructing this future external validation set is provided in the Appendix.
We hope PlantXpert will serve as a foundation for future research on evidence-grounded multimodal reasoning, model adaptation, and scientifically reliable evaluation in plant science and agriculture.

\section*{Declaration of generative AI use}
During the preparation of this work, the authors used OpenAI GPT-5 in order to improve the readability and language of the manuscript. 
After using this tool, the authors reviewed and edited the content as needed and take full responsibility for the content of the published article.

\section*{CRediT authorship contribution statement}
\textbf{Yu Wu}: Conceptualization, Methodology, Data curation, Visualization, Formal analysis, Writing – original draft, Review \& editing.
\textbf{Guangzeng Han}: Conceptualization, Methodology, Data curation, Visualization, Formal analysis, Writing - original draft, Review \& editing.
\textbf{Ibra Niang Niang}: Data curation, Visualization, Formal analysis, Writing - original draft.
\textbf{Francia Ravelombola}: Data curation, Review \& editing.
\textbf{Maiara Oliveira}: Data curation, Review \& editing.
\textbf{Jason Davis}: Review \& editing, Resources.
\textbf{Dong Chen}: Review \& editing, Resources.
\textbf{Feng Lin}: Review \& editing, Supervision, Resources, Project administration, Funding acquisition.
\textbf{Xiaolei Huang}: Conceptualization, Methodology, Writing \& review \& editing, Supervision, Resources, Project administration, Funding acquisition.

\section*{Declaration of competing interest}
The authors declare that they have no known competing financial interests or personal relationships that could have appeared to influence the work reported in this paper.

\section*{Data availability}
Data will be made available on request.

\section*{Acknowledgment}

The project was partially supported by the National Science Foundation under award IIS-2549203.
We thank Zihan Xu and Fnu Abhishek for assisting data annotations.
We thank for the computing resources provided by the iTiger GPU cluster~\citep{sharif_2025_ITIGER} supported by the NSF MRI program (CNS-2318210).
We would also thank for the ITS at the University of Memphis to partially support the HPC operations.

\appendix

\section{Appendix: Source Data Summary}
We present the summary of retrieved source data in Table~\ref{tab:retrieval_stats}.
\begin{table}[H]
\centering
\caption{Summary of retrieved articles and figures.}
\label{tab:retrieval_stats}

\begin{tabular}{l l c c}

\textbf{Crop} & \textbf{Keywords} & \textbf{Articles} & \textbf{Figures} \\
\midrule
Soybean & Disease assessment, pest control, weed management, crop yield & 1,685 & 305 \\
Cotton  & Disease assessment, pest control, weed management, crop yield & 793 & 80 \\
Rice  & Disease assessment, pest control, weed management, crop yield & 800 & 80 \\
Corn  & Disease assessment, pest control, weed management, crop yield & 800 & 80 \\

\end{tabular}
\end{table}

\section{Appendix: Prompts For Data Generation}
\label{appendx:prompt}
\lstset{
    backgroundcolor=\color[RGB]{245,245,245},
    breaklines=true,
    breakindent=0pt,
    basicstyle=\ttfamily\small,
    frame=single,           
    xleftmargin=0pt,        
    xrightmargin=0pt,
    framexleftmargin=0pt,
    framexrightmargin=0pt
}
\begin{lstlisting}
    Create 8 CHALLENGING Agriculture VQA questions that would be difficult for current advanced AI models.\n
    IMPORTANT: The questions will be answered using ONLY the image - no caption or context will be provided to models during testing.\n
    Reference info (for question generation only):
    - Caption: {figure_data['caption']}
    - Context: {' '.join(figure_data['related_paragraphs'])}
    - Crop: {paper_info['crop_type']}
    - Figure type: {figure_data.get('figure_type', 'unknown')}
    
    Assign each question:
    - one expert_field from:
      ["Plant Health & Pathology", "Plant Biology & Physiology", "Crop Science & Agronomy", "Remote Sensing & Imaging", "Entomology & Weed Science", "Others"]
    - one primary_challenge from:
      ["Reasoning", "Visual", "Quantitative", "Others"]
    
    Do NOT force a balanced distribution across categories.
    Instead, infer the most appropriate domain and challenge type for each question based on the image and the reference information.
    The overall mix of the 8 questions should emerge naturally from the figure content. If one domain or challenge type is much more strongly supported by the figure, it is acceptable for multiple questions to share that label.
    

    Design questions that can be answered from visual information alone, but require:
    1. **Multi-step reasoning**: Some questions require 3-4 logical steps from visual evidence
    2. **Model weaknesses**: Some target spatial reasoning, counting, negation handling
    3. **PhD-level expertise**: Deep agricultural knowledge to interpret visual symptoms/features
    4. **High-quality distractors**: Sophisticated wrong answers that look plausible \n
    Create 8 DIVERSE questions that are answerable from the image alone but extremely challenging.\n
    JSON format:
    {{
      "questions": [
        {{
          "question": "Your question (answerable from image only)...",
          "options": {{
            "A": "Option A",
            "B": "Option B", 
            "C": "Option C",
            "D": "Option D"
          }},
          "correct_answer": "C",
          "primary_challenge": "Quantitative & Statistical Estimation",
          "expert_field": "Plant Biology & Physiology",
          "difficulty_explanation": "Why this is challenging",
          "visual_evidence": "What visual clues lead to the answer"
        }}
      ]
    }}
\end{lstlisting}

\section{Appendix: Candidate Sources for Future External Validation}

We provide a list of candidate publications to construct a future external validation set in Table~\ref{tab:external_sources}. 
These publications are not included in the current PlantXpert benchmark and are provided as potential sources for future evaluation.

\begin{longtable}{l p{8.2cm} l}
\caption{Candidate publications for future external validation.}
\label{tab:external_sources} \\

\textbf{Crop} & \textbf{Publication} & \textbf{Visual Evidence} \\
\midrule
\endfirsthead

\textbf{Crop} & \textbf{Publication} & \textbf{Visual Evidence} \\
\midrule
\endhead

Soybean &
Development of an adaptive sprayer control system for UAV-based spot spraying in soybean. &
UAV imagery \\

Soybean &
A Data-Centric SPGT-ConvTransformer Framework for UAV-Based Soybean Disease Classification. &
Disease imagery \\

Soybean &
Texture- and threshold-based image segmentation for quantifying Asian soybean rust severity in digital images. &
Leaf disease images \\

Soybean &
Optimization and Experimental Evaluation of a Deep Learning-Based Target Spraying Device for Weed Control in Soybean Fields. &
Field RGB images \\

Soybean &
RFDAF-Net: a novel region-specific feature decoupling and adaptive fusion network for field soybean disease identification in precision agriculture. &
Field disease images \\

\midrule

Cotton &
Detection and Precision Application Path Planning for Cotton Spider Mite Based on UAV Multispectral Remote Sensing. &
UAV multispectral imagery \\

Cotton &
Assessing nitrogen levels and crop yield prediction in cotton using unmanned aerial vehicle-based multispectral remote sensing. &
UAV multispectral imagery \\

Cotton &
Field-based deep learning classification of cotton and weeds for machine-vision-assisted intra-row weed management. &
Field RGB images \\

Cotton &
An attention enhanced CNN ensemble for interpretable and accurate cotton leaf disease classification. &
Leaf disease images \\

Cotton &
Development of a Robotic Weed Puller for Precision Management of Palmer Amaranth in Cotton. &
Field robot imagery \\

\midrule

Rice &
BLB\_TVI: A hyperspectral triangular index for satellite-scale characterization of rice bacterial leaf blight severity. &
Hyperspectral imagery \\

Rice &
Field-scale rice yield prediction using UAV imagery and machine learning in a developing country context. &
UAV imagery \\

Rice &
YOLOv12-RSLW: An Efficient Detection and Severity Grading Framework for Rice False Smut via Count-Area Calibration. &
Panicle images \\

Rice &
Enhancing rice breeding efficiency through semi-supervised detection and segmentation of panicles and leaves. &
Panicle and leaf images \\

Rice &
A Field-Deployable Visual Monitoring Device for Measuring Nocturnal Phototactic Rhythm of Rice Pests. &
Pest images \\

\midrule

Corn &
Estimation of the three-dimensional spatial distribution of maize leaf chlorophyll through fusion of spectral, textural, and structural features from multimodal remote sensing. &
Multimodal imagery \\

Corn &
Identification of the optimal phenological periods for summer maize yield prediction using UAV-based multispectral data. &
UAV multispectral imagery \\

Corn &
Real-time hyperspectral decision support for precision herbicide management in maize: Optimizing herbicide efficacy and plant response. &
Hyperspectral imagery \\

Corn &
Enhanced corn leaf disease detection using sharpness-aware minimization optimized CNNs. &
Leaf disease images \\

Corn &
Multimodal deep learning model for real-time severity assessment of maize stalk rot integrating visual and soil sensing data. &
Stalk imagery \\

\bottomrule
\end{longtable}





\bibliographystyle{elsarticle-harv}


\bibliography{bib1}

@article{li_figure_2019,
    author = {Li, Pengyuan and Jiang, Xiangying and Shatkay, Hagit},
    title = {Figure and caption extraction from biomedical documents},
    journal = {Bioinformatics},
    volume = {35},
    number = {21},
    pages = {4381-4388},
    year = {2019},
    month = {04},
    issn = {1367-4803},
    doi = {10.1093/bioinformatics/btz228},
    url ={https://academic.oup.com/bioinformatics/article/35/21/4381/5428177}
}

@INPROCEEDINGS{Arshad_WACV_2025,
  author={Arshad, Muhammad Arbab and Jubery, Talukder Zaki and Roy, Tirtho and Nassiri, Rim and Singh, Asheesh K. and Singh, Arti and Hegde, Chinmay and Ganapathysubramanian, Baskar and Balu, Aditya and Krishnamurthy, Adarsh and Sarkar, Soumik},
  booktitle={2025 IEEE/CVF Winter Conference on Applications of Computer Vision (WACV)}, 
  title={Leveraging Vision Language Models for Specialized Agricultural Tasks}, 
  year={2025},
  pages={6320-6329},
  doi={10.1109/WACV61041.2025.00616},
  url = {https://ieeexplore.ieee.org/document/10943968}
  }

@InProceedings{Shinoda_ICCV_2025,
    author    = {Shinoda, Risa and Inoue, Nakamasa and Kataoka, Hirokatsu and Onishi, Masaki and Ushiku, Yoshitaka},
    title     = {AgroBench: Vision-Language Model Benchmark in Agriculture},
    booktitle = {Proceedings of the IEEE/CVF International Conference on Computer Vision (ICCV)},
    month     = {October},
    year      = {2025},
    pages     = {7634-7644},
    url = {https://openaccess.thecvf.com/content/ICCV2025/html/Shinoda_AgroBench_Vision-Language_Model_Benchmark_in_Agriculture_ICCV_2025_paper.html}
}

@article{wang_artificial_2025,
	address = {Switzerland},
	title = {Artificial intelligence in plant science: from image-based phenotyping to yield and trait prediction.},
	volume = {16},
	issn = {1664-462X},
	doi = {10.3389/fpls.2025.1732979},
    url = {https://www.frontiersin.org/journals/plant-science/articles/10.3389/fpls.2025.1732979/full},
	language = {eng},
	journal = {Frontiers in plant science},
	author = {Wang, Tong and Tong, Ran and Xu, Ting and Li, Yue and Chen, Yonghao},
	year = {2025},
	pages = {1732979}
}

@inproceedings{zhou_agribench_2025,
	address = {Cham},
	title = {AgriBench: A Hierarchical Agriculture Benchmark for Multimodal Large Language Models},
	isbn = {978-3-031-91835-3},
	booktitle = {Computer Vision – ECCV 2024 Workshops},
	publisher = {Springer Nature Switzerland},
	author = {Zhou, Yutong and Ryo, Masahiro},
	editor = {Del Bue, Alessio and Canton, Cristian and Pont-Tuset, Jordi and Tommasi, Tatiana},
	year = {2025},
	pages = {207--223},
    doi = {10.1007/978-3-031-91835-3_14},
    url = {https://link.springer.com/chapter/10.1007/978-3-031-91835-3_14}
}

@inproceedings{liu_multimodal_2025,
	address = {Cham},
	title = {A Multimodal Benchmark Dataset and Model for Crop Disease Diagnosis},
	isbn = {978-3-031-73016-0},
	booktitle = {Computer Vision – ECCV 2024},
	publisher = {Springer Nature Switzerland},
	author = {Liu, Xiang and Liu, Zhaoxiang and Hu, Huan and Chen, Zezhou and Wang, Kohou and Wang, Kai and Lian, Shiguo},
	editor = {Leonardis, Aleš and Ricci, Elisa and Roth, Stefan and Russakovsky, Olga and Sattler, Torsten and Varol, Gül},
	year = {2025},
	pages = {157--170},
    doi = {10.1007/978-3-031-73016-0_10},
    url = {https://link.springer.com/chapter/10.1007/978-3-031-73016-0_10}
}

@article{mir_high-throughput_2019,
	title = {High-throughput phenotyping for crop improvement in the genomics era},
	volume = {282},
	issn = {01689452},
	doi = {10.1016/j.plantsci.2019.01.007},
    url = {https://www.sciencedirect.com/science/article/abs/pii/S0168945218305752},
	language = {en},
	journal = {Plant Science},
	author = {Mir, Reyazul Rouf and Reynolds, Mathew and Pinto, Francisco and Khan, Mohd Anwar and Bhat, Mohd Ashraf},
	month = may,
	year = {2019},
	pages = {60--72}
}

@misc{bommasani_opportunities_2022,
	title = {On the Opportunities and Risks of Foundation Models},
    url = {https://arxiv.org/abs/2108.07258},
	language = {en},
	publisher = {arXiv preprint},
	author = {Bommasani, Rishi and Hudson, Drew A. and Adeli, Ehsan and Altman, Russ and Arora, Simran and Arx, Sydney von and Bernstein, Michael S. and Bohg, Jeannette and Bosselut, Antoine and Brunskill, Emma and Brynjolfsson, Erik and Buch, Shyamal and Card, Dallas and Castellon, Rodrigo and Chatterji, Niladri and Chen, Annie and Creel, Kathleen and Davis, Jared Quincy and Demszky, Dora and Donahue, Chris and Doumbouya, Moussa and Durmus, Esin and Ermon, Stefano and Etchemendy, John and Ethayarajh, Kawin and Fei-Fei, Li and Finn, Chelsea and Gale, Trevor and Gillespie, Lauren and Goel, Karan and Goodman, Noah and Grossman, Shelby and Guha, Neel and Hashimoto, Tatsunori and Henderson, Peter and Hewitt, John and Ho, Daniel E. and Hong, Jenny and Hsu, Kyle and Huang, Jing and Icard, Thomas and Jain, Saahil and Jurafsky, Dan and Kalluri, Pratyusha and Karamcheti, Siddharth and Keeling, Geoff and Khani, Fereshte and Khattab, Omar and Koh, Pang Wei and Krass, Mark and Krishna, Ranjay and Kuditipudi, Rohith and Kumar, Ananya and Ladhak, Faisal and Lee, Mina and Lee, Tony and Leskovec, Jure and Levent, Isabelle and Li, Xiang Lisa and Li, Xuechen and Ma, Tengyu and Malik, Ali and Manning, Christopher D. and Mirchandani, Suvir and Mitchell, Eric and Munyikwa, Zanele and Nair, Suraj and Narayan, Avanika and Narayanan, Deepak and Newman, Ben and Nie, Allen and Niebles, Juan Carlos and Nilforoshan, Hamed and Nyarko, Julian and Ogut, Giray and Orr, Laurel and Papadimitriou, Isabel and Park, Joon Sung and Piech, Chris and Portelance, Eva and Potts, Christopher and Raghunathan, Aditi and Reich, Rob and Ren, Hongyu and Rong, Frieda and Roohani, Yusuf and Ruiz, Camilo and Ryan, Jack and Ré, Christopher and Sadigh, Dorsa and Sagawa, Shiori and Santhanam, Keshav and Shih, Andy and Srinivasan, Krishnan and Tamkin, Alex and Taori, Rohan and Thomas, Armin W. and Tramèr, Florian and Wang, Rose E. and Wang, William and Wu, Bohan and Wu, Jiajun and Wu, Yuhuai and Xie, Sang Michael and Yasunaga, Michihiro and You, Jiaxuan and Zaharia, Matei and Zhang, Michael and Zhang, Tianyi and Zhang, Xikun and Zhang, Yuhui and Zheng, Lucia and Zhou, Kaitlyn and Liang, Percy},
	month = {jul},
	year = {2022}
}

@article{williamson_data_2023,
	title = {Data management challenges for artificial intelligence in plant and agricultural research},
	language = {en},
	author = {Williamson, Hugh F and Brettschneider, Julia and Caccamo, Mario and Davey, Robert P and Goble, Carole and Kersey, Paul J and May, Sean and Morris, Richard J and Ostler, Richard and Pridmore, Tony and Rawlings, Chris and Studholme, David and Tsaftaris, Sotirios A and Leonelli, Sabina},
	year = {2021},
    volume = {10},
    pages = {324},
    journal = {F1000Research},
    doi = {10.12688/f1000research.52204.2},
    url = {https://f1000research.com/articles/10-324}
}

@article{mohanty_using_2016,
	title = {Using Deep Learning for Image-Based Plant Disease Detection},
	volume = {7},
	issn = {1664-462X},
	doi = {10.3389/fpls.2016.01419},
    url = {https://www.frontiersin.org/journals/plant-science/articles/10.3389/fpls.2016.01419/full},
	language = {en},
	journal = {Frontiers in Plant Science},
	author = {Mohanty, Sharada P. and Hughes, David P. and Salathé, Marcel},
	month = {sep},
	year = {2016},
	pages = {1419}
}

@article{meshram_fruitnet_2022,
	title = {FruitNet: Indian fruits image dataset with quality for machine learning applications},
	volume = {40},
	issn = {23523409},
	shorttitle = {FruitNet},
	doi = {10.1016/j.dib.2021.107686},
    url = {https://www.sciencedirect.com/science/article/pii/S2352340921009616},
	language = {en},
	journal = {Data in Brief},
	author = {Meshram, Vishal and Patil, Kailas},
	month = feb,
	year = {2022},
	pages = {107686}
}

@inproceedings{singh_plantdoc_2020,
    author = {Singh, Davinder and Jain, Naman and Jain, Pranjali and Kayal, Pratik and Kumawat, Sudhakar and Batra, Nipun},
    title = {PlantDoc: A Dataset for Visual Plant Disease Detection},
    year = {2020},
    isbn = {9781450377386},
    publisher = {Association for Computing Machinery},
    address = {New York, NY, USA},
    url = {https://dl.acm.org/doi/abs/10.1145/3371158.3371196},
    doi = {10.1145/3371158.3371196},
    booktitle = {Proceedings of the 7th ACM IKDD CoDS and 25th COMAD},
    pages = {249–253},
    numpages = {5},
    location = {Hyderabad, India},
    series = {CoDS COMAD 2020}
}

@techreport{bai_qwen3_2025,
    archivePrefix = {arXiv preprint},
    arxivId = {2511.21631},
    author = {Bai, Shuai and Cai, Yuxuan and Chen, Ruizhe and Chen, Keqin and Chen, Xionghui and Cheng, Zesen and Deng, Lianghao and Ding, Wei and Gao, Chang and Ge, Chunjiang and Ge, Wenbin and Guo, Zhifang and Huang, Qidong and Huang, Jie and Huang, Fei and Hui, Binyuan and Jiang, Shutong and Li, Zhaohai and Li, Mingsheng and Li, Mei and Li, Kaixin and Lin, Zicheng and Lin, Junyang and Liu, Xuejing and Liu, Jiawei and Liu, Chenglong and Liu, Yang and Liu, Dayiheng and Liu, Shixuan and Lu, Dunjie and Luo, Ruilin and Lv, Chenxu and Men, Rui and Meng, Lingchen and Ren, Xuancheng and Ren, Xingzhang and Song, Sibo and Sun, Yuchong and Tang, Jun and Tu, Jianhong and Wan, Jianqiang and Wang, Peng and Wang, Pengfei and Wang, Qiuyue and Wang, Yuxuan and Xie, Tianbao and Xu, Yiheng and Xu, Haiyang and Xu, Jin and Yang, Zhibo and Yang, Mingkun and Yang, Jianxin and Yang, An and Yu, Bowen and Zhang, Fei and Zhang, Hang and Zhang, Xi and Zheng, Bo and Zhong, Humen and Zhou, Jingren and Zhou, Fan and Zhou, Jing and Zhu, Yuanzhi and Zhu, Ke},
    eprint = {2511.21631},
    institution = {Alibaba Cloud},
    month = {nov},
    title = {{Qwen3-VL Technical Report}},
    url = {http://arxiv.org/abs/2511.21631},
    year = {2025}
}

@techreport{gemmateam_gemma3technicalreport_2025,
      title={Gemma 3 Technical Report}, 
      author={Gemma Team and Aishwarya Kamath and Johan Ferret and Shreya Pathak and Nino Vieillard and Ramona Merhej and Sarah Perrin and Tatiana Matejovicova and Alexandre Ramé and Morgane Rivière and Louis Rouillard and Thomas Mesnard and Geoffrey Cideron and Jean-bastien Grill and Sabela Ramos and Edouard Yvinec and Michelle Casbon and Etienne Pot and Ivo Penchev and Gaël Liu and Francesco Visin and Kathleen Kenealy and Lucas Beyer and Xiaohai Zhai and Anton Tsitsulin and Robert Busa-Fekete and Alex Feng and Noveen Sachdeva and Benjamin Coleman and Yi Gao and Basil Mustafa and Iain Barr and Emilio Parisotto and David Tian and Matan Eyal and Colin Cherry and Jan-Thorsten Peter and Danila Sinopalnikov and Surya Bhupatiraju and Rishabh Agarwal and Mehran Kazemi and Dan Malkin and Ravin Kumar and David Vilar and Idan Brusilovsky and Jiaming Luo and Andreas Steiner and Abe Friesen and Abhanshu Sharma and Abheesht Sharma and Adi Mayrav Gilady and Adrian Goedeckemeyer and Alaa Saade and Alex Feng and Alexander Kolesnikov and Alexei Bendebury and Alvin Abdagic and Amit Vadi and András György and André Susano Pinto and Anil Das and Ankur Bapna and Antoine Miech and Antoine Yang and Antonia Paterson and Ashish Shenoy and Ayan Chakrabarti and Bilal Piot and Bo Wu and Bobak Shahriari and Bryce Petrini and Charlie Chen and Charline Le Lan and Christopher A. Choquette-Choo and CJ Carey and Cormac Brick and Daniel Deutsch and Danielle Eisenbud and Dee Cattle and Derek Cheng and Dimitris Paparas and Divyashree Shivakumar Sreepathihalli and Doug Reid and Dustin Tran and Dustin Zelle and Eric Noland and Erwin Huizenga and Eugene Kharitonov and Frederick Liu and Gagik Amirkhanyan and Glenn Cameron and Hadi Hashemi and Hanna Klimczak-Plucińska and Harman Singh and Harsh Mehta and Harshal Tushar Lehri and Hussein Hazimeh and Ian Ballantyne and Idan Szpektor and Ivan Nardini and Jean Pouget-Abadie and Jetha Chan and Joe Stanton and John Wieting and Jonathan Lai and Jordi Orbay and Joseph Fernandez and Josh Newlan and Ju-yeong Ji and Jyotinder Singh and Kat Black and Kathy Yu and Kevin Hui and Kiran Vodrahalli and Klaus Greff and Linhai Qiu and Marcella Valentine and Marina Coelho and Marvin Ritter and Matt Hoffman and Matthew Watson and Mayank Chaturvedi and Michael Moynihan and Min Ma and Nabila Babar and Natasha Noy and Nathan Byrd and Nick Roy and Nikola Momchev and Nilay Chauhan and Noveen Sachdeva and Oskar Bunyan and Pankil Botarda and Paul Caron and Paul Kishan Rubenstein and Phil Culliton and Philipp Schmid and Pier Giuseppe Sessa and Pingmei Xu and Piotr Stanczyk and Pouya Tafti and Rakesh Shivanna and Renjie Wu and Renke Pan and Reza Rokni and Rob Willoughby and Rohith Vallu and Ryan Mullins and Sammy Jerome and Sara Smoot and Sertan Girgin and Shariq Iqbal and Shashir Reddy and Shruti Sheth and Siim Põder and Sijal Bhatnagar and Sindhu Raghuram Panyam and Sivan Eiger and Susan Zhang and Tianqi Liu and Trevor Yacovone and Tyler Liechty and Uday Kalra and Utku Evci and Vedant Misra and Vincent Roseberry and Vlad Feinberg and Vlad Kolesnikov and Woohyun Han and Woosuk Kwon and Xi Chen and Yinlam Chow and Yuvein Zhu and Zichuan Wei and Zoltan Egyed and Victor Cotruta and Minh Giang and Phoebe Kirk and Anand Rao and Kat Black and Nabila Babar and Jessica Lo and Erica Moreira and Luiz Gustavo Martins and Omar Sanseviero and Lucas Gonzalez and Zach Gleicher and Tris Warkentin and Vahab Mirrokni and Evan Senter and Eli Collins and Joelle Barral and Zoubin Ghahramani and Raia Hadsell and Yossi Matias and D. Sculley and Slav Petrov and Noah Fiedel and Noam Shazeer and Oriol Vinyals and Jeff Dean and Demis Hassabis and Koray Kavukcuoglu and Clement Farabet and Elena Buchatskaya and Jean-Baptiste Alayrac and Rohan Anil and Dmitry and Lepikhin and Sebastian Borgeaud and Olivier Bachem and Armand Joulin and Alek Andreev and Cassidy Hardin and Robert Dadashi and Léonard Hussenot},
      institution = {Google},
      year={2025},
      archivePrefix={arXiv preprint},
      primaryClass={cs.CL},
      url={https://arxiv.org/abs/2503.19786}, 
}

@techreport{wang_internvl3_2025,
  title={Internvl3. 5: Advancing open-source multimodal models in versatility, reasoning, and efficiency},
  author={Wang, Weiyun and Gao, Zhangwei and Gu, Lixin and Pu, Hengjun and Cui, Long and Wei, Xingguang and Liu, Zhaoyang and Jing, Linglin and Ye, Shenglong and Shao, Jie and others},
  journal={arXiv preprint},
  url={https://arxiv.org/abs/2508.18265}, 
  institution = {Shanghai AI Laboratory},
  year={2025}
}

@misc{liu_llavanext_2024,
    title={LLaVA-NeXT: Improved reasoning, OCR, and world knowledge},
    url={https://llava-vl.github.io/blog/2024-01-30-llava-next/},
    author={Liu, Haotian and Li, Chunyuan and Li, Yuheng and Li, Bo and Zhang, Yuanhan and Shen, Sheng and Lee, Yong Jae},
    month={January},
    year={2024}
}

@techreport{team_granite_2025,
  title={Granite Vision: a lightweight, open-source multimodal model for enterprise Intelligence},
  author={Team, Granite Vision and Karlinsky, Leonid and Arbelle, Assaf and Daniels, Abraham and Nassar, Ahmed and Alfassi, Amit and Wu, Bo and Schwartz, Eli and Joshi, Dhiraj and Kondic, Jovana and others},
  journal={arXiv preprint},
  url={https://arxiv.org/abs/2502.09927}, 
  institution={IBM Research},
  year={2025}
}

@article{Weyler_PhenoBench_2024,
  title={PhenoBench: A Large Dataset and Benchmarks for Semantic Image Interpretation in the Agricultural Domain}, 
  author={Weyler, Jan and Magistri, Federico and Marks, Elias and Chong, Yue Linn and Sodano, Matteo and Roggiolani, Gianmarco and Chebrolu, Nived and Stachniss, Cyrill and Behley, Jens},
  journal={IEEE Transactions on Pattern Analysis and Machine Intelligence}, 
  year={2024},
  volume={46},
  number={12},
  pages={9583-9594},
  doi={10.1109/TPAMI.2024.3419548}
}

@inproceedings{hu_lora_2022,
  title={Lo{RA}: Low-Rank Adaptation of Large Language Models},
  author={Hu, Edward J and Shen, Yelong and Wallis, Phillip and Allen-Zhu, Zeyuan and Li, Yuanzhi and Wang, Shean and Wang, Liang and Chen, Weizhu and others},
  booktitle={International Conference on Learning Representations},
  year={2022},
  pages={13},
  url={https://openreview.net/forum?id=nZeVKeeFYf9}
}

@incollection{waghmare_cotton_2022,
	address = {Singapore},
	title = {Cotton Breeding},
	isbn = {978-981-16-9257-4},
	doi = {10.1007/978-981-16-9257-4_11},
    url = {https://link.springer.com/chapter/10.1007/978-981-16-9257-4_11},
	booktitle = {Fundamentals of Field Crop Breeding},
	publisher = {Springer Nature Singapore},
	author = {Waghmare, Vijay N.},
	editor = {Yadava, Devendra Kumar and Dikshit, Harsh Kumar and Mishra, Gyan Prakash and Tripathi, Shailesh},
	year = {2022},
	pages = {609--676},
}

@incollection{anderson_soybean_2019,
	address = {Cham},
	title = {Soybean [Glycine max (L.) Merr.] Breeding: History, Improvement, Production and Future Opportunities},
	isbn = {978-3-030-23400-3},
	doi = {10.1007/978-3-030-23400-3_12},
    url = {https://link.springer.com/chapter/10.1007/978-3-030-23400-3_12},
	booktitle = {Advances in Plant Breeding Strategies: Legumes: Volume 7},
	publisher = {Springer International Publishing},
	author = {Anderson, Edwin J. and Ali, Md Liakat and Beavis, William D. and Chen, Pengyin and Clemente, Tom Elmo and Diers, Brian W. and Graef, George L. and Grassini, Patricio and Hyten, David L. and McHale, Leah K. and Nelson, Randall L. and Parrott, Wayne A. and Patil, Gunvant B. and Stupar, Robert M. and Tilmon, Kelley J.},
	editor = {Al-Khayri, Jameel M. and Jain, Shri Mohan and Johnson, Dennis V.},
	year = {2019},
	pages = {431--516},
}

@article{wei_plantseg_2024,
    author = {Wei, Tianqi and Chen, Zhi and Yu, Xin and Chapman, Scott and Melloy, Paul and Huang, Zi},
    doi = {10.1038/s41597-025-06513-4},
    issn = {2052-4463},
    journal = {Scientific Data},
    month = {feb},
    number = {1},
    pages = {205},
    title = {{A Large-Scale In-the-wild Dataset for Plant Disease Segmentation}},
    url = {https://www.nature.com/articles/s41597-025-06513-4},
    volume = {13},
    year = {2026}
}

@techreport{sharif_2025_ITIGER,
      title={Cultivating Multidisciplinary Research and Education on GPU Infrastructure for Mid-South Institutions at the University of Memphis: Practice and Challenge}, 
      author={Sharif, Mayira and Han, Guangzeng and Liu, Weisi and Huang, Xiaolei},
      year={2025},
      eprint={2504.14786},
      institution={University of Memphis},
      primaryClass={cs.DC},
      url={https://arxiv.org/abs/2504.14786}, 
}

@misc{anthropic_claude_2026,
	title = {Claude Sonnet 5.5 System Card},
	url = {https://www.anthropic.com/claude-sonnet-5-5-system-card},
	language = {en},
	publisher = {Anthropic},
	author = {{Anthropic}},
	year = {2026},
	month = sep
}

@misc{singh_openai_2025,
	title = {OpenAI GPT-5 System Card},
	url = {http://arxiv.org/abs/2601.03267},
	language = {en},
	urldate = {2026-03-27},
	publisher = {arXiv},
	author = {Singh, Aaditya and Fry, Adam and Perelman, Adam and Tart, Adam and Ganesh, Adi and El-Kishky, Ahmed and McLaughlin, Aidan and Low, Aiden and Ostrow, A. J. and Ananthram, Akhila and Nathan, Akshay and Luo, Alan and Helyar, Alec and Madry, Aleksander and Efremov, Aleksandr and Spyra, Aleksandra and Baker-Whitcomb, Alex and Beutel, Alex and Karpenko, Alex and Makelov, Alex and Neitz, Alex and Wei, Alex and Barr, Alexandra and Kirchmeyer, Alexandre and Ivanov, Alexey and Christakis, Alexi and Gillespie, Alistair and Tam, Allison and Bennett, Ally and Wan, Alvin and Huang, Alyssa and Sandjideh, Amy McDonald and Yang, Amy and Kumar, Ananya and Saraiva, Andre and Vallone, Andrea and Gheorghe, Andrei and Garcia, Andres Garcia and Braunstein, Andrew and Liu, Andrew and Schmidt, Andrew and Mereskin, Andrey and Mishchenko, Andrey and Applebaum, Andy and Rogerson, Andy and Rajan, Ann and Wei, Annie and Kotha, Anoop and Srivastava, Anubha and Agrawal, Anushree and Vijayvergiya, Arun and Tyra, Ashley and Nair, Ashvin and Nayak, Avi and Eggers, Ben and Ji, Bessie and Hoover, Beth and Chen, Bill and Chen, Blair and Barak, Boaz and Minaiev, Borys and Hao, Botao and Baker, Bowen and Lightcap, Brad and McKinzie, Brandon and Wang, Brandon and Quinn, Brendan and Fioca, Brian and Hsu, Brian and Yang, Brian and Yu, Brian and Zhang, Brian and Brenner, Brittany and Zetino, Callie Riggins and Raymond, Cameron and Lugaresi, Camillo and Paz, Carolina and Hudson, Cary and Whitney, Cedric and Li, Chak and Chen, Charles and Cole, Charlotte and Voss, Chelsea and Ding, Chen and Shen, Chen and Huang, Chengdu and Colby, Chris and Hallacy, Chris and Koch, Chris and Lu, Chris and Kaplan, Christina and Kim, Christina and Minott-Henriques, C. J. and Frey, Cliff and Yu, Cody and Czarnecki, Coley and Reid, Colin and Wei, Colin and Decareaux, Cory and Scheau, Cristina and Zhang, Cyril and Forbes, Cyrus and Tang, Da and Goldberg, Dakota and Roberts, Dan and Palmie, Dana and Kappler, Daniel and Levine, Daniel and Wright, Daniel and Leo, Dave and Lin, David and Robinson, David and Grabb, Declan and Chen, Derek and Lim, Derek and Salama, Derek and Bhattacharjee, Dibya and Tsipras, Dimitris and Li, Dinghua and Yu, Dingli and Strouse, D. J. and Williams, Drew and Hunn, Dylan and Bayes, Ed and Arbus, Edwin and Akyurek, Ekin and Le, Elaine Ya and Widmann, Elana and Yani, Eli and Proehl, Elizabeth and Sert, Enis and Cheung, Enoch and Schwartz, Eri and Han, Eric and Jiang, Eric and Mitchell, Eric and Sigler, Eric and Wallace, Eric and Ritter, Erik and Kavanaugh, Erin and Mays, Evan and Nikishin, Evgenii and Li, Fangyuan and Such, Felipe Petroski and Peres, Filipe de Avila Belbute and Raso, Filippo and Bekerman, Florent and Tsimpourlas, Foivos and Chantzis, Fotis and Song, Francis and Zhang, Francis and Raila, Gaby and McGrath, Garrett and Briggs, Gary and Yang, Gary and Parascandolo, Giambattista and Chabot, Gildas and Kim, Grace and Zhao, Grace and Valiant, Gregory and Leclerc, Guillaume and Salman, Hadi and Wang, Hanson and Sheng, Hao and Jiang, Haoming and Wang, Haoyu and Jin, Haozhun and Sikchi, Harshit and Schmidt, Heather and Aspegren, Henry and Chen, Honglin and Qiu, Huida and Lightman, Hunter and Covert, Ian and Kivlichan, Ian and Silber, Ian and Sohl, Ian and Hammoud, Ibrahim and Clavera, Ignasi and Lan, Ikai and Akkaya, Ilge and Kostrikov, Ilya and Kofman, Irina and Etinger, Isak and Singal, Ishaan and Hehir, Jackie and Huh, Jacob and Pan, Jacqueline and Wilczynski, Jake and Pachocki, Jakub and Lee, James and Quinn, James and Kiros, Jamie and Kalra, Janvi and Samaroo, Jasmyn and Wang, Jason and Wolfe, Jason and Chen, Jay and Wang, Jay and Harb, Jean and Han, Jeffrey and Wang, Jeffrey and Zhao, Jennifer and Chen, Jeremy and Yang, Jerene and Tworek, Jerry and Chand, Jesse and Landon, Jessica and Liang, Jessica and Lin, Ji and Liu, Jiancheng and Wang, Jianfeng and Tang, Jie and Yin, Jihan and Jang, Joanne and Morris, Joel and Flynn, Joey and Ferstad, Johannes and Heidecke, Johannes and Fishbein, John and Hallman, John and Grant, Jonah and Chien, Jonathan and Gordon, Jonathan and Park, Jongsoo and Liss, Jordan and Kraaijeveld, Jos and Guay, Joseph and Mo, Joseph and Lawson, Josh and McGrath, Josh and Vendrow, Joshua and Jiao, Joy and Lee, Julian and Steele, Julie and Wang, Julie and Mao, Junhua and Chen, Kai and Hayashi, Kai and Xiao, Kai and Salahi, Kamyar and Wu, Kan and Sekhri, Karan and Sharma, Karan and Singhal, Karan and Li, Karen and Nguyen, Kenny and Gu-Lemberg, Keren and King, Kevin and Liu, Kevin and Stone, Kevin and Yu, Kevin and Ying, Kristen and Georgiev, Kristian and Lim, Kristie and Tirumala, Kushal and Miller, Kyle and Ahmad, Lama and Lv, Larry and Clare, Laura and Fauconnet, Laurance and Itow, Lauren and Yang, Lauren and Romaniuk, Laurentia and Anise, Leah and Byron, Lee and Pathak, Leher and Maksin, Leon and Lo, Leyan and Ho, Leyton and Jing, Li and Wu, Liang and Xiong, Liang and Mamitsuka, Lien and Yang, Lin and McCallum, Lindsay and Held, Lindsey and Bourgeois, Liz and Engstrom, Logan and Kuhn, Lorenz and Feuvrier, Louis and Zhang, Lu and Switzer, Lucas and Kondraciuk, Lukas and Kaiser, Lukasz and Joglekar, Manas and Singh, Mandeep and Shah, Mandip and Stratta, Manuka and Williams, Marcus and Chen, Mark and Sun, Mark and Cayton, Marselus and Li, Martin and Zhang, Marvin and Aljubeh, Marwan and Nichols, Matt and Haines, Matthew and Schwarzer, Max and Gupta, Mayank and Shah, Meghan and Huang, Melody and Dong, Meng and Wang, Mengqing and Glaese, Mia and Carroll, Micah and Lampe, Michael and Malek, Michael and Sharman, Michael and Zhang, Michael and Wang, Michele and Pokrass, Michelle and Florian, Mihai and Pavlov, Mikhail and Wang, Miles and Chen, Ming and Wang, Mingxuan and Feng, Minnia and Bavarian, Mo and Lin, Molly and Abdool, Moose and Rohaninejad, Mostafa and Soto, Nacho and Staudacher, Natalie and LaFontaine, Natan and Marwell, Nathan and Liu, Nelson and Preston, Nick and Turley, Nick and Ansman, Nicklas and Blades, Nicole and Pancha, Nikil and Mikhaylin, Nikita and Felix, Niko and Handa, Nikunj and Rai, Nishant and Keskar, Nitish and Brown, Noam and Nachum, Ofir and Boiko, Oleg and Murk, Oleg and Watkins, Olivia and Gleeson, Oona and Mishkin, Pamela and Lesiewicz, Patryk and Baltescu, Paul and Belov, Pavel and Zhokhov, Peter and Pronin, Philip and Guo, Phillip and Thacker, Phoebe and Liu, Qi and Yuan, Qiming and Liu, Qinghua and Dias, Rachel and Puckett, Rachel and Arora, Rahul and Mullapudi, Ravi Teja and Gaon, Raz and Miyara, Reah and Song, Rennie and Aggarwal, Rishabh and Marsan, R. J. and Yemiru, Robel and Xiong, Robert and Kshirsagar, Rohan and Nuttall, Rohan and Tsiupa, Roman and Eldan, Ronen and Wang, Rose and James, Roshan and Ziv, Roy and Shu, Rui and Nigmatullin, Ruslan and Jain, Saachi and Talaie, Saam and Altman, Sam and Arnesen, Sam and Toizer, Sam and Toyer, Sam and Miserendino, Samuel and Agarwal, Sandhini and Yoo, Sarah and Heon, Savannah and Ethersmith, Scott and Grove, Sean and Taylor, Sean and Bubeck, Sebastien and Banesiu, Sever and Amdo, Shaokyi and Zhao, Shengjia and Wu, Sherwin and Santurkar, Shibani and Zhao, Shiyu and Chaudhuri, Shraman Ray and Krishnaswamy, Shreyas and Shuaiqi and Xia and Cheng, Shuyang and Anadkat, Shyamal and Fishman, Simón Posada and Tobin, Simon and Fu, Siyuan and Jain, Somay and Mei, Song and Egoian, Sonya and Kim, Spencer and Golden, Spug and Mah, S. Q. and Lin, Steph and Imm, Stephen and Sharpe, Steve and Yadlowsky, Steve and Choudhry, Sulman and Eum, Sungwon and Sanjeev, Suvansh and Khan, Tabarak and Stramer, Tal and Wang, Tao and Xin, Tao and Gogineni, Tarun and Christianson, Taya and Sanders, Ted and Patwardhan, Tejal and Degry, Thomas and Shadwell, Thomas and Fu, Tianfu and Gao, Tianshi and Garipov, Timur and Sriskandarajah, Tina and Sherbakov, Toki and Kaftan, Tomer and Hiratsuka, Tomo and Wang, Tongzhou and Song, Tony and Zhao, Tony and Peterson, Troy and Kharitonov, Val and Chernova, Victoria and Kosaraju, Vineet and Kuo, Vishal and Pong, Vitchyr and Verma, Vivek and Petrov, Vlad and Jiang, Wanning and Zhang, Weixing and Zhou, Wenda and Xie, Wenlei and Zhan, Wenting and McCabe, Wes and DePue, Will and Ellsworth, Will and Bain, Wulfie and Thompson, Wyatt and Chen, Xiangning and Qi, Xiangyu and Xiang, Xin and Shi, Xinwei and Dubois, Yann and Yu, Yaodong and Khakbaz, Yara and Wu, Yifan and Qian, Yilei and Lee, Yin Tat and Chen, Yinbo and Zhang, Yizhen and Xiong, Yizhong and Tian, Yonglong and Cha, Young and Bai, Yu and Yang, Yu and Yuan, Yuan and Li, Yuanzhi and Zhang, Yufeng and Yang, Yuguang and Jin, Yujia and Jiang, Yun and Wang, Yunyun and Wang, Yushi and Liu, Yutian and Stubenvoll, Zach and Dou, Zehao and Wu, Zheng and Wang, Zhigang},
	month = {dec},
	year = {2025},
	note = {arXiv:2601.03267 [cs]}
}

@article{Sapkota_multimodal_2025,
  author={Sapkota, Ranjan and Qureshi, Rizwan and Usman Hadi, Muhammad and Zohaib Hassan, Syed and Sadak, Ferhat and Shoman, Maged and Sajjad, Muhammad and Ali Dharejo, Fayaz and Paudel, Achyut and Li, Jiajia and Meng, Zhichao and Shutske, John and Karkee, Manoj},
  journal={IEEE Transactions on Automation Science and Engineering}, 
  title={Multi-Modal LLMs in Agriculture: A Comprehensive Review}, 
  year={2025},
  volume={22},
  number={},
  pages={22510-22540},
  doi={10.1109/TASE.2025.3612154},
  URL={https://ieeexplore.ieee.org/document/11173627}}

@article{li_foundation_2024,
	title = {Foundation models in smart agriculture: Basics, opportunities, and challenges},
	volume = {222},
	issn = {0168-1699},
	url = {https://www.sciencedirect.com/science/article/pii/S016816992400423X},
	doi = {https://doi.org/10.1016/j.compag.2024.109032},
	journal = {Computers and Electronics in Agriculture},
	author = {Li, Jiajia and Xu, Mingle and Xiang, Lirong and Chen, Dong and Zhuang, Weichao and Yin, Xunyuan and Li, Zhaojian},
	year = {2024},
	pages = {109032}
}

@article{yan_agrieval_2026,
	title = {AgriEval: A Comprehensive Chinese Agricultural Benchmark for Large Language Models},
	volume = {40},
	url = {https://ojs.aaai.org/index.php/AAAI/article/view/40716},
	doi = {10.1609/aaai.v40i40.40716},
	number = {40},
	journal = {Proceedings of the AAAI Conference on Artificial Intelligence},
	author = {Yan, Lian and Wang, Haotian and Tang, Chen and Liu, Haifeng and Sun, Tianyang and Liu, Liangliang and Guan, Yi and Jiang, Jingchi},
	month = {mar},
	year = {2026},
	pages = {34205--34213},
}

@inproceedings{awais_agrogpt_2025,
	address = {Tucson, AZ, USA},
	title = {AgroGPT : Efficient Agricultural Vision-Language Model with Expert Tuning},
	isbn = {979-8-3315-1083-1},
	shorttitle = {AgroGPT},
	url = {https://ieeexplore.ieee.org/document/10944186/},
	doi = {10.1109/WACV61041.2025.00555},
	language = {en},
	urldate = {2026-02-20},
	booktitle = {2025 {IEEE}/{CVF} {Winter} {Conference} on {Applications} of {Computer} {Vision} ({WACV})},
	publisher = {IEEE},
	author = {Awais, Muhammad and Salem Abdulla Alharthi, Ali Husain and Kumar, Amandeep and Cholakkal, Hisham and Anwer, Rao Muhammad},
	month = {feb},
	year = {2025},
	pages = {5687--5696},
}

@misc{wang_agri-llava_2024,
	title = {Agri-LLaVA: Knowledge-Infused Large Multimodal Assistant on Agricultural Pests and Diseases},
	language = {en},
    year={2024},
	author = {Wang, Liqiong and Jin, Teng and Yang, Jinyu and Leonardis, Alesˇ and Wang, Fangyi and Zheng, Feng},
    publisher = {arxiv preprint},
    primaryClass={cs.CV},
    url={https://arxiv.org/abs/2412.02158}
}

@misc{yang_agrigpt_2025,
	title = {AgriGPT: a Large Language Model Ecosystem for Agriculture},
	shorttitle = {AgriGPT},
	url = {http://arxiv.org/abs/2508.08632},
	language = {en},
	publisher = {arXiv preprint},
	author = {Yang, Bo and Zhang, Yu and Feng, Lanfei and Chen, Yunkui and Zhang, Jianyu and Xu, Xiao and Aierken, Nueraili and Li, Yurui and Chen, Yuxuan and Yang, Guijun and He, Yong and Huang, Runhe and Li, Shijian},
	month = {aug},
	year = {2025},
}

@misc{quoc_leafnet_2026,
	title = {LeafNet: A Large-Scale Dataset and Comprehensive Benchmark for Foundational Vision-Language Understanding of Plant Diseases},
	shorttitle = {LeafNet},
	url = {http://arxiv.org/abs/2602.13662},
	language = {en},
	publisher = {arXiv preprint},
	author = {Quoc, Khang Nguyen and Dao, Phuong D. and Quach, Luyl-Da},
	month = feb,
	year = {2026},
}

@inproceedings{wei_chain_2022,
    author = {Wei, Jason and Wang, Xuezhi and Schuurmans, Dale and Bosma, Maarten and ichter, brian and Xia, Fei and Chi, Ed and Le, Quoc V and Zhou, Denny},
    booktitle = {Advances in Neural Information Processing Systems},
    editor = {S. Koyejo and S. Mohamed and A. Agarwal and D. Belgrave and K. Cho and A. Oh},
    pages = {24824--24837},
    publisher = {Curran Associates, Inc.},
    title = {Chain-of-Thought Prompting Elicits Reasoning in Large Language Models},
    url = {https://proceedings.neurips.cc/paper_files/paper/2022/file/9d5609613524ecf4f15af0f7b31abca4-Paper-Conference.pdf},
    volume = {35},
    year = {2022}
}

@misc{Label,
  title={{Label Studio}: Data labeling software},
  url={https://github.com/heartexlabs/label-studio},
  note={Open source software available from https://github.com/heartexlabs/label-studio},
  author={
    Maxim Tkachenko and
    Mikhail Malyuk and
    Andrey Holmanyuk and
    Nikolai Liubimov},
  year={2020-2022},
}

@inproceedings{zheng-llamafactory-2024,
    title = {{L}lama{F}actory: Unified Efficient Fine-Tuning of 100+ Language Models},
    author = {Zheng, Yaowei  and Zhang, Richong  and Zhang, Junhao  and Ye, Yanhan  and Luo, Zheyan},
    editor = {Cao, Yixin  and Feng, Yang  and Xiong, Deyi},
    booktitle = {Proceedings of the 62nd Annual Meeting of the Association for Computational Linguistics (Volume 3: System Demonstrations)},
    month = aug,
    year = {2024},
    address = {Bangkok, Thailand},
    publisher = {Association for Computational Linguistics},
    url = {https://aclanthology.org/2024.acl-demos.38/},
    doi = {10.18653/v1/2024.acl-demos.38},
    pages = {400--410},
}

@article{liang2025dynamic,
  title={Dynamic text prompt joint multimodal features for accurate plant disease image captioning},
  author={Liang, Fangfang and Huang, Zilong and Wang, Wenjian and He, Zhenxue and En, Qing},
  journal={The Visual Computer},
  volume={41},
  number={8},
  pages={5405--5419},
  year={2025},
  publisher={Springer Berlin Heidelberg Berlin/Heidelberg}
}

@article{zhang2024cmmmu,
  title={Cmmmu: A chinese massive multi-discipline multimodal understanding benchmark},
  author={Zhang, Ge and Du, Xinrun and Chen, Bei and Liang, Yiming and Luo, Tongxu and Zheng, Tianyu and Zhu, Kang and Cheng, Yuyang and Xu, Chunpu and Guo, Shuyue and others},
  journal={arXiv preprint arXiv:2401.11944},
  year={2024}
}

@inproceedings{yue2024mmmu,
  title={Mmmu: A massive multi-discipline multimodal understanding and reasoning benchmark for expert agi},
  author={Yue, Xiang and Ni, Yuansheng and Zhang, Kai and Zheng, Tianyu and Liu, Ruoqi and Zhang, Ge and Stevens, Samuel and Jiang, Dongfu and Ren, Weiming and Sun, Yuxuan and others},
  booktitle={Proceedings of the IEEE/CVF conference on computer vision and pattern recognition},
  pages={9556--9567},
  year={2024}
}

@article{quoc2026leafnet,
  title={LeafNet: A Large-Scale Dataset and Comprehensive Benchmark for Foundational Vision-Language Understanding of Plant Diseases},
  author={Quoc, Khang Nguyen and Dao, Phuong D and Quach, Luyl-Da},
  journal={arXiv preprint arXiv:2602.13662},
  year={2026}
}

@article{mohidem_rice_2022,
	title = {Rice for Food Security: Revisiting Its Production, Diversity, Rice Milling Process and Nutrient Content},
	volume = {12},
	issn = {2077-0472},
	url = {https://www.mdpi.com/2077-0472/12/6/741},
	doi = {10.3390/agriculture12060741},
    number = {6},
	journal = {Agriculture},
	author = {Mohidem, Nur Atikah and Hashim, Norhashila and Shamsudin, Rosnah and Che Man, Hasfalina},
	year = {2022},
}

@incollection{GARCIALARA_corn_2019,
title = {Chapter 1 - Corn History and Culture},
editor = {Sergio O. Serna-Saldivar},
booktitle = {Corn (Third Edition)},
publisher = {AACC International Press},
edition = {Third Edition},
address = {Oxford},
pages = {1-18},
year = {2019},
isbn = {978-0-12-811971-6},
doi = {https://doi.org/10.1016/B978-0-12-811971-6.00001-2},
url = {https://www.sciencedirect.com/science/article/pii/B9780128119716000012},
author = {Silverio García-Lara and Sergio O. Serna-Saldivar},
}

\end{document}